\documentclass[journal]{IEEEtran}
\usepackage[noadjust]{cite}

\usepackage[dvips]{graphicx}
\usepackage{multirow,multicol}

\usepackage[tbtags]{amsmath}
\usepackage{amsbsy}
\usepackage{amssymb}
\usepackage{amsfonts}
\usepackage{bbm}

\usepackage{graphicx}
\usepackage[caption=false]{subfig}
\usepackage{pstricks}
\usepackage{pst-node}
\usepackage{pstricks-add}
\usepackage{url}
\usepackage{textcomp}
\usepackage{algorithmic}
\usepackage{algorithm}
\usepackage{balance}

\usepackage{array}

{\begin{list}               
    {$\bullet$ \hfill}{
        \setlength{\leftmargin}{\parindent}
        \setlength{\parsep}{0.04\baselineskip}
        \setlength{\itemsep}{0.5\parsep}
        \setlength{\labelwidth}{\leftmargin}
        \setlength{\labelsep}{0em}}
    }
{\end{list}}

\providecommand{\eref}[1]{\eqref{#1}}  
\providecommand{\cref}[1]{Chapter~\ref{#1}}
\providecommand{\sref}[1]{Section~\ref{#1}}
\providecommand{\fref}[1]{Fig.~\ref{#1}}

\providecommand{\R}{\ensuremath{\mathbb{R}}}

\providecommand{\Z}{\ensuremath{\mathbb{Z}}}

\renewcommand{\vec}[1]{\ensuremath{\boldsymbol{#1}}}
\providecommand{\mat}[1]{\ensuremath{\boldsymbol{#1}}}

\providecommand{\calR}{\mathcal{R}}

\providecommand{\mA}{\mat{A}}
\providecommand{\mB}{\mat{B}}
\providecommand{\mC}{\mat{C}}

\providecommand{\mE}{\mat{E}}

\providecommand{\mP}{\mat{P}}
\providecommand{\mQ}{\mat{Q}}

\providecommand{\mS}{\mat{S}}

\providecommand{\mU}{\mat{U}}
\providecommand{\mV}{\mat{V}}

\providecommand{\mX}{\mat{X}}

\providecommand{\vh}{\vec{h}}

\providecommand{\vl}{\vec{l}}

\providecommand{\vn}{\vec{n}}

\providecommand{\vp}{\vec{p}}

\providecommand{\vr}{\vec{r}}
\providecommand{\vs}{\vec{s}}

\providecommand{\vv}{\vec{v}}

\providecommand{\vx}{\vec{x}}
\providecommand{\vy}{\vec{y}}

\providecommand{\mSigma}{\mat{\Sigma}}

\providecommand{\vphi}{\vec{\phi}}

\providecommand{\vpsi}{\vec{\psi}}

\newcommand{\defequal}{\mathop{\overset{\mbox{\tiny{def}}}{=}}}

\DeclareGraphicsExtensions{.eps}

\begin{document}

\title{Joint Defogging and Demosaicking}

\author{Yeejin~Lee,~\IEEEmembership{Student~Member,~IEEE,} 
        Keigo~Hirakawa,~\IEEEmembership{Senior~Member,~IEEE,}
        and~Truong~Q.~Nguyen,~\IEEEmembership{Fellow,~IEEE}} 

\maketitle

\begin{abstract}
Image defogging is a technique used extensively for enhancing visual quality of images in bad weather condition. Even though defogging algorithms have been well studied, defogging performance is degraded by demosaicking artifacts and sensor noise amplification in distant scenes. In order to improve visual quality of restored images, we propose a novel approach to perform defogging and demosaicking simultaneously. We conclude that better defogging performance with fewer artifacts can be achieved when a defogging algorithm is combined with a demosaicking algorithm simultaneously. We also demonstrate that the proposed joint algorithm has the benefit of suppressing noise amplification in distant scene. In addition, we validate our theoretical analysis and observations for both synthesized datasets with ground truth fog-free images and natural scene datasets captured in a raw format.  
\end{abstract}


\section{Introduction}
\IEEEPARstart{T}{he} scene radiance captured in bad weather conditions, such as fog and haze, is attenuated as the light travels towards the camera because of atmospheric scattering~\cite{Middleton52, McCartney76}. When combined with the environment light, the acquired image appears faint---image contrast, color saturation, and visibility progressively worsen with the increased scene distance. As these degradations negatively impact feature extraction, it is often necessary to improve the image quality before the captured images are usable in computer vision applications such as surveillance, security, traffic monitoring, and driver assistance. 

Defogging or dehazing refers to a post-capture technique to restore the contrast of images captured under such conditions by reversing the effects of atmospheric scattering. Besides the challenges of estimating the scene depth (needed for parameterizing the severity of the fog), it is well known that the defogging process is prone to noise amplification~\cite{Schechner07}. The fact that defogging can also amplify artifacts caused by various stages of the digital camera processing pipeline has received very limited attention in the literature, however. Schechner~\cite{Schechner07} proposed a method to recover images while suppressing noise that is dependent on distance. Tarel~\cite{Tarel09} proposed a way to smooth the restored images to reduce the compression artifacts that become visible after defogging. Gibson {\it{et al.}}~\cite{Gibson12} showed that fewer artifacts and better coding efficiency are achieved when a defogging algorithm is applied before compression. 

There are other ways in which the digital camera processing pipeline interacts with defogging in a nontrivial manner. Specifically, raw data captured by an image sensor are intrinsically unrecognizable to the human eye. The digital camera processing pipeline refers to a set of image processing algorithms aimed at transforming the captured raw sensor data into images faithfully representing the scene that the photographer saw. A typical pipeline is comprised of defective sensor pixel removal, demosaicking (also known as color filter array interpolation), color correction, gamma correction, noise suppression, and data formatting (see Figure \ref{fig:DSC}). Any imperfections or artifacts introduced by these tasks can potentially be amplified by the subsequent defogging process.

In this paper, we focus on the influence of demosaicking artifacts and noise on defogging. Our analysis in Section \ref{sec:background} of the three way interaction between image denoising, demosaicking, and defogging suggests that the problem is prevalent, regardless of which order in which these operations are performed. To this end, we propose an alternative digital camera processing pipeline framework based on combining demosaicking, denoising, and defogging. In Section \ref{sec:algorithm}, we first develop a patch-based defogging algorithm that leverages total least squares (TLS) regression to improve the robustness of defogging process to noise when fog is severe. We then extend this method to perform defogging and demosaicking processes simultaneously. The joint optimization of defogging and demosaicking has the advantage that demosaicking filter coefficients are tuned to minimize post-defogging residual error. As a side note, combining defogging and demosaicking also implies that these steps take place prior to gamma correction. We contrast this to the post-processing approach to defogging, where due to the gamma correction step in the digital camera processing pipeline, the pixel values no longer scale linearly to the scene radiance. This contributes to an additional source of uncertainty in the defogging problem since fog is a physical phenomenon best described in terms of light intensity~\cite{Schechner07, Tarel09, Lee15}. This fact also points to the challenges of assessing the defogging performance in simulation using images in canonical formats such as sRGB, where the pixel values do not directly correspond to any radiometric values. Therefore, we validate our proposed method using real raw image sensor data of actual foggy scenes  (Section \ref{sec:results}).

The rest of the paper is organized as follows. In Section~\ref{sec:background} we provide background on existing defogging approaches. A detailed analysis of the influence of digital camera processing pipeline on defogging is also presented. Next, an alternative digital camera processing pipeline framework is introduced in Section~\ref{sec:algorithm} and the proposed joint defogging and demosaicking algorithm is presented. Then, experimental results are presented in Section~\ref{sec:results}. Finally, we conclude with a summary of key observations in Section~\ref{sec:conclusion}.

\section{Background} \label{sec:background}
\subsection{Defogging}
Atmospheric scattering is caused by small particles such as water droplets suspended in air that diffuse the light propagating through it, away from the camera. In a homogeneous medium, the spectral scene radiance attenuates exponentially with respect to the path length $d$ and the attenuation coefficient $\beta$ for a specific wavelength $\lambda$. As described in the Beer-Lambert-Bouguer law~\cite{McCartney76}, the spectral transmittance $t$ (the fraction of radiation actually transmitted without scattering) at the pixel location $\left(i,j\right)$ is
\begin{equation}
t \left( i, j, \lambda \right) = e^{-\beta\left(i, j, \lambda\right)d\left( i, j\right)}.
\end{equation}
In this paper, we approximate the scattering coefficient $\beta\left(i,j,\lambda\right)=\beta$ to be known, spatially invariant, and constant with respect to wavelengths $\lambda$~\cite{Middleton52}---i.e.~$t(i,j,\lambda)=t(i,j)$ is a function dependent on the distance $d(i,j)$ only. The same suspended particles also scatter \emph{airlight} (or environment light such as sun and sky) a portion of which is reflected towards the camera. The blending of the attenuated scene radiance and the airlight is described by the Koschmieder’s law as follows~\cite{Middleton52,McCartney76, Narasimhan99}:
\begin{equation}\label{eq:FoggyModel}
\vy\left( i, j\right) = t\left(i, j \right) \vx\left( i, j \right)+\left( 1- t\left(i, j\right)\right)\vl_{a},
\end{equation}
where $\vx:\Z^2\to \R^{3}$ is the scatter-free scene radiance, $\vy:\Z^2\to\R^{3}$ is the observed light, and  $\vl_{a} =\lim_{d\to \infty} \vy \in \R^{3}$ is airlight observed as a path radiance at infinite scene distance. We conclude that the contribution from the airlight increases with the path length $d(i,j)$ while the scene radiance attenuates, causing a distant object to appear lighter and faint.

Since \eqref{eq:FoggyModel} is reversible if transmission and airlight are known, fog-free image $\vx(i,j)$ is recoverable from $\vy(i,j)$ as follows:
\begin{equation} \label{eq:RecoveryModel}
\vx\left( i,j\right) = \frac{\vy\left(i,j\right)-\vl_{a}}{\max\left( t\left(i,j\right), \epsilon\right)} + \vl_{a} \:,
\end{equation}
where $\epsilon$ is a small constant for numerical stability. In practice, defogging is complicated by the fact that transmission $t(i,j)$ and airlight $\vl_{a}$ in \eqref{eq:FoggyModel} are unknown. 

Various methods to learn these parameters based on physical models have been proposed. Narasimhan~\cite{Narasimhan99, Narasimhan03} extracted depth edges from the images acquired in different atmospheric scattering conditions. Atmospheric transmission can be estimated from airlight measured in different degrees of polarization~\cite{Schechner01, Shwartz06, Schechner07, Fang14}. These multi-image methods are sensitive to weather conditions, camera positions, and camera parameters. In applications such as in navigation where the camera is not stationary, multiple image acquisition is less feasible and even impractical.

Alternatively, parameters of \eqref{eq:FoggyModel} may be inferred from a single observation by exploiting physical scene structures~\cite{Oakley02, Narasimhan99, Narasimhan03}. Cozman~\cite{Cozman97} developed a technique to extract depth ratio between two points located on the same object in a single image. Oakley enhanced degraded contrast by using environmental parameters and temporal information with known scene geometry in~\cite{Oakley02}. Kopf {\it et al.} also used known 3D geometrical models to estimate more accurate transmission along with foggy images~\cite{Kopf08}. In vehicle navigation, a method in \cite{Hautiere07} estimated the scene structure from predicting the vanishing point of a road. Improved defogging performance by leveraging depth cues in stereo imaging is reported in~\cite{Caraffa12} and \cite{Lee14}.  

Among the statistical approaches to a single-image defogging, He {\it et al}.~\cite{He09} estimated atmospheric scattering using statistics of fog-free outdoor images, denoted as the dark channel prior~(DCP). Fattal~\cite{Fattal08} recovered albedo and scene structure by assuming that they are statistically independent. Similarly, a technique in \cite{Kratz09} factorized foggy images into two statistically independent latent layers of scene albedo and depth in a factorial MRF, which was solved with the help of the expectation-maximization~(EM) algorithm. Caraffal~\cite{Caraffa12} extended the MRF model from a single image defogging to a stereo image pair. Tan~\cite{Tan08} optimized a recovered scene by maximizing local contrast and by smoothing airlight using a Gaussian Markov random field~(GMRF). More recently, Wang~\cite{Wang14} combined the multiscale depth fusion with the Laplacian-MRF to recover the depth map.

There have been additional efforts to enforce spatial consistency in the estimated transmission maps. Tarel~\cite{Tarel09} and Gibson~\cite{Gibson13} leveraged edge-preserving filters to determine the depth and fog density transitions in a scene. A variational method was used in~\cite{Schechner07} to regularize spatially varying transmission. 

\begin{figure*}[!t]\begin{center} 
\includegraphics[scale=0.5]{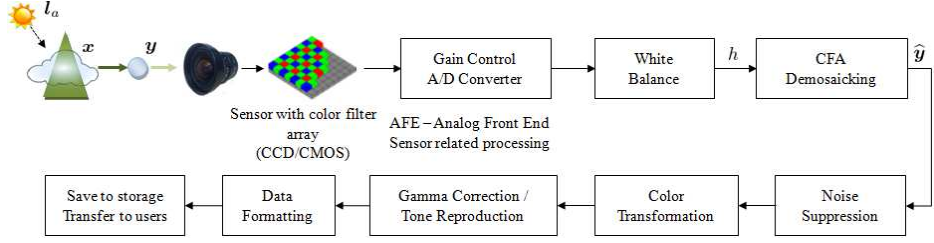}
\caption{An example of digital camera processing pipeline. The pipeline shows how the raw foggy data proceeds from the formation to the viewer. The composition and order of components differ from camera manufacturers.} \label{fig:DSC}
\end{center}\end{figure*}  

\subsection{Image Sensor}\label{sec:AnalysisNoise}
Noise is present in all images captured by image sensors. Two most common sources of noise in image sensors are fixed pattern noise and random noise. Fixed pattern noise stems from variabilities in pixel sensors or analog-to-digital converters, and can be reduced to an extent by improving manufacturing tolerance. This type of hardware noise is ``repeatable''---meaning its impact can also be reduced significantly by proper calibration; hence, fixed pattern noise is not one of our primary concerns in our work. On the other hand, random noise refers to the non-repeatable noise that cannot be eliminated solely by hardware improvements~\cite{Jin14}. Shot noise stemming from the stochasticity of the photon arrival process is noticeable in poorly illuminated scenes. This is also a concern for foggy images where the signal is attenuated significantly. 

The noise in the sensor observation $\vs(i,j)$ at location $(i,j)$ is classically modeled as a combination of Poisson distribution parameterized by the signal intensity $\vy(i,j)$ and the readout noise $\vn(i,j)$, as follows~\cite{Jin14}:
\begin{align} \label{eq:NoiseModel}
\begin{split}
\vs(i,j) &= \vp(i,j) + \vn(i,j)\\
\vp(i,j) &\sim \mathcal{P}(\vy(i,j)),
\end{split}
\end{align} 
where the variance of the photon count $\vp(i,j)$ scales linearly with the intensity of the foggy image $\vy(i,j)$; the additive readout noise $\vn(i,j)$ is often modeled as additive constant-rate Poisson or additive white Gaussian with signal-independent noise variance $\sigma^2$. The overall noise variance of $\vs(i,j)$ is 
\begin{align} 
\begin{split}
\operatorname{var}(\vs(i,j))&=y(i,j)+\sigma^2\\
&=t(i,j)\vx(i,j)+(1-t(i,j))\vl_a+\sigma^2,
\end{split}
\end{align}
where we have abused the notation to mean
\begin{align}
\operatorname{var}(\vs(i,j))=
\begin{bmatrix}
\operatorname{var}(s_1(i,j))\\
\operatorname{var}(s_2(i,j))\\
\operatorname{var}(s_3(i,j))
\end{bmatrix}.
\end{align}
In the presence of heavy fog (i.e.~large $\beta$) or at long scene distance (i.e.~large $d(i,j)$), the noise is dominated by the airlight $\vl_a$ and the readout noise:
\begin{align}\label{eq:NoiseAmplification}
\lim_{t\to 0} \operatorname{var}(\vs(i,j))=\vl_a+\sigma^2.
\end{align}
Replacing $\vy(i,j)$ with its noisy version in \eqref{eq:NoiseModel}, defogging procedure in \eqref{eq:RecoveryModel} suffers from noise amplification:
\begin{align}
\operatorname{var}\left(\frac{\vs(i,j)-\vl_a}{t(i,j)}+\vl_a\right) = \frac{\operatorname{var}(\vs(i,j))}{\max\left(t(i,j),\epsilon\right)^2},
\end{align}
which quickly approaches $\frac{\vl_a+\sigma^2}{t(i,j)^2}$ as $t(i,j)\to 0$. We conclude that defogged images are dominated by signal independent additive noise that is amplified drastically with the scene distance; even a negligible measurement uncertainty can significantly degrade the restored images~\cite{Schechner07}. As such, assuming that distance is sufficiently large and invoking normal approximation, we make the following simplifying approximation for the remainder of this paper:
\begin{align}
\vs(i,j) \sim\mathcal{N}(\vy(i,j),\operatorname{diag}(\vl_a+\sigma^2)),
\end{align}
where $\operatorname{diag}$ is an operation to form a square diagonal matrix with the elements of a vector on the main diagonal.

As a side note, we point out that the techniques for learning noise parameters from a single image typically fail with foggy images. Specifically, fog degrades image contrast, making it difficult to identify homogeneous patches within an image where the noise variance estimates are most reliable. With the increased likelihood of misclassifying the inhomogeneous regions of the images as homogeneous, the estimated noise variance is typically larger than the actual noise power. The inflated noise variance consequently oversmoothes the scene when applying denoising methods to foggy images~(see Fig.~\ref{fig:DefoggedWithAlphas}). 

\subsection{Demosaicking}\label{sec:AnalysisDemosaicking}
In a conventional camera, image sensor is equipped with color filter array (CFA)---a spatial multiplexing of red, green, and blue filters over the array of pixel sensors. Given a color image $\vx(i,j)$, the sensor captures CFA sampled values $g:\Z^2\to\R$, as follows:
\begin{equation}\begin{array}{ccl}\label{cfa1}
g(i,j) & = &\vphi(i,j)^{T}\vx(i,j),
\end{array}\end{equation}  
where $\vphi:\Z^2\to \{0,1\}^3$ is a sampling lattice (e.g.~$\vphi(i,j) = \left[ 1\:0\:0\right]^{T}$ denotes a red sample at pixel location $(i,j)$).
Under the influence of a dense fog, a CFA-sampled sensor data of a foggy image instead takes the form:
\begin{equation}\begin{array}{ccl}\label{cfa2}
h(i,j) & = &\vphi(i,j)^{T}\vs(i,j)\vspace{0.1cm}\\
  & = & \vphi(i,j)^{T} \left[ t(i,j)\vx +\left( 1-t(i,j)\right)\vl_{a}+\vn(i,j)\right]\vspace{0.1cm}\\
  & = & t(i,j)g(i,j)+\left( 1-t(i,j)\right) l_{a}+n(i,j),
\end{array}\end{equation}  
where $\vn(i,j)\sim\mathcal{N}(0,\operatorname{diag}(\vl_a)+\sigma^2)$, $l_{a}=\vphi(i,j)^{T}\vl_{a}$ is a constant for a properly white-balanced image (i.e.~$\vl_a=[l_a,l_a,l_a]^T$), and $n(i,j)=\vphi(i,j)^T\vn(i,j)$.

The demosaicking process interpolates the CFA image in \eqref{cfa1} and reconstructs the color image. Typically, the goal is to design an operator $\vpsi\{\cdot\}$ to recover the signal $\vx(i,j)$ from the observation $g:\Z^2\to\R$:
\begin{align}
\widehat{\vx}(i,j)&=\vpsi\{g\}(i,j).
\end{align}
 Under the influence of a dense fog, however, we have instead
\begin{align}\label{eq:demosaickedfoggy}
\begin{split}
\widehat{\vy}(i,j)=&\vpsi\{ h\}(i,j)\\
=&\vpsi\{t\cdot g+\left( 1-t\right) l_{a}+n\}(i,j)\\
\approx&\vpsi\{t\cdot g+n\}(i,j) + \left( 1-t(i,j)\right) \vl_{a},
\end{split}
\end{align}
where the approximation in \eqref{eq:demosaickedfoggy} holds if $t(i,j)$ is spatially slowly varying---most demosaicking methods are invariant to offset by a constant $k$ (i.e.~$\vpsi\{g+k\}\approx\vpsi\{g\}+k$) because the edge magnitudes are not influenced. On the other hand, the demosaicking error in $\widetilde{\vy}(i,j)=\vpsi\{t\cdot g+n\}(i,j)-t\cdot \vx(i,j)$ stems from the fact that edge magnitudes are attenuated by $t\in[0,1]$. This increases the risks of misclassifying the inhomogeneous regions of the image as homogeneous regions and consequently yielding a demosaicked image $\vpsi\{t\cdot g\}(i,j)$ that is considerably smoother than $\vx(i,j)$, or increasing the risks of the image details falling below the noise floor.

Assuming that the demosaicked image $\widehat{\vy}(i,j)$ is propagated through the defogging process in \eqref{eq:RecoveryModel}, then the interpolation errors are exacerbated by the subsequent defogging process, as shown below: 
\begin{align} 
\begin{split}
& \frac{\widehat{\vy}\left(i,j\right)-\vl_{a}}{\max\left( t\left(i,j\right), \epsilon\right)} + \vl_{a} \\
&\approx  \frac{t\cdot \vx(i,j) + \widetilde{\vy}(i,j) + \left( 1-t(i,j)\right) \vl_{a}-\vl_{a}}{\max\left( t\left(i,j\right), \epsilon\right)} + \vl_{a}\\
&= \vx(i,j)+\frac{\widetilde{\vy}(i,j)}{\max\left( t\left(i,j\right), \epsilon\right)},
\end{split}
\end{align}
because $t(i,j)\in[0,1]$. In addition, the defogging process is complicated by the fact that the transmission is unknown. We conclude that demosaicking error $\widetilde{\vy}(i,j)$ alters the recovered scene radiance that negatively impacts the estimation of transmission and airlight. See \fref{fig:DefoggedWithAlphas}.

\section{Proposed Algorithm}\label{sec:algorithm}

The aforementioned analysis suggests that demosaicking artifacts $\widetilde{\vy}(i,j)$ and sensor noise $\vn(i,j)$ are amplified by the defogging algorithm. We therefore propose an alternative digital camera processing pipeline that jointly solves the defogging and demosaicking problems. The proposed method is designed to minimize the post-defogging demosaicking artifacts and sensor noise. We address the problem in Section~\ref{sec:TLSProbState} and formulate it using the total least squares method in~\ref{sec:ProbForm}. Section~\ref{sec:ProbSolving} solves the joint problem and Section~\ref{sec:Implementation} develops the implementation details of the algorithm. 

\subsection{Problem Statement}\label{sec:TLSProbState}

Given a noisy, Bayer CFA sampled data $h(i,j)$~\cite{Bayer76} of a foggy image $\vs(i,j)$ as described by \eqref{eq:FoggyModel}, \eqref{eq:NoiseModel}, and \eqref{cfa2}, the goal of joint defogging and demosaicking is to compute the estimate $\widehat{\vx}(i,j)$ of the latent fog-free color image $\vx(i,j)$. The physical parameters such as transmission $t(i,j)$ and airlight $\vl_a$ in \eqref{eq:FoggyModel} are also unknown \emph{a priori}. 

We leverage a demosaicking method in~\cite{Hirakawa06} that takes advantage of the scale-invariant pixel correlation structure within an image patch~\cite{Ruderman93, Khashabi14} and cross color correlation~\cite{Gunturk02, Hirakawa05ahd}. Specifically, we estimate the fog-free $k$th color component of the desired color image $\vx(i,j)=[x_1(i,j),x_2(i,j),x_3(i,j)]^T$ as a linear combination of pixels in $h$:
\begin{align}\label{eq:estimate}
\widehat{x}_k(i,j)= \sum_{m,n} w(m,n)(h(i+m,j+n)-\beta(m,n))+\gamma(i,j),
\end{align}
where $w,\beta,\gamma$ are the parameters of the estimator. For clarity, we let $k=1$ (i.e.~we are estimating $x_k(i,j)$ that is a red sample) in the subsequent presentation---the results obtained here generalizes to the estimation of green and blue pixels. Recalling the fact that the spatially highpass components of red, green, and blue color images are highly correlated~\cite{Gunturk02, Hirakawa05ahd}, the contribution of green and blue color pixels in $\vh$ to the estimation of a red pixel $x_k(i,j)$ should be limited to spatially highpass components~\cite{Gunturk02, Hirakawa05ahd}. One way to accomplish this is to ensure that the weights $w$ corresponding to green and blue color pixels should add up to zero so that spatially lowpass components are attenuated. To this end, we rewrite $w$ as 
\begin{align}\label{eq:constraint}
w(m,n)=\sum_{\ell=1}^L \alpha_{\ell} c_{\ell}(m,n),
\end{align}
where $\{c_1(m,n),\dots,c_L(m,n)\}$ is a set of predetermined basis vectors whose green and blue pixel weights sum to zero, and the parameters of the ``constrained'' estimator are now $\alpha_{\ell},\beta,\gamma$. Note that \eqref{eq:estimate} is well-suited for the joint demosaicking and defogging problem involving incomplete color information of CFA pattern and unknown variables in fog-free images. When $h(i,j)$ is a red sample, \eqref{eq:estimate} represents a noise-suppressed image defogging method; when $h(i,j)$ is a green or blue sample, \eqref{eq:estimate} is a demosaicking method that estimates the missing fog-free red sample.

The constraints in \eqref{eq:constraint} also give rise to the following substitution:
\begin{align}\label{eq:substitution}
\widehat{x}_k(i,j) \approx \sum_{m,n} w(m,n)(s_k(i+m,j+n)-\beta'(m,n))+\gamma(i,j),
\end{align}
that is, the estimation of fog-free red pixel value $\widehat{x}_k(i,j)$ from a complete foggy red image $s_k:\Z^2\to\R$. Though this is justified because the spatially highpass component of green and blue pixels can be substituted by the highpass of red, \eqref{eq:substitution} is not implementable in practice because not all of $s_k$ is observed. Nevertheless,  we draw on the autoregressive principals to learn $\alpha_{\ell}$ that yield weights $w(i,j)$ and the estimate $\widehat{x}_k(i,j)$ which best approximate the desiderata $x_k(i,j)$~\cite{Ruderman93}. Specifically, let $(i',j')$ be a pixel location near $(i,j)$ such that $s_k(i',j')=h(i',j')$ (i.e.~red pixel $(i',j')$ closest to $(i,j)$). Then
\begin{align}\label{eq:autoregression}
\begin{split}
&\widehat{x}'_k(i',j') \\
&= \sum_{m,n} w(m,n)(s_k(i'+2m,j'+2n)-\beta'(m,n))+\gamma(i',j'),
\end{split}
\end{align}
where $\widehat{x}'_k(i',j')$ is another estimate of $x_k(i',j')$, and we used the fact that $s_k(i'+2m,j'+2n)=h(i'+2m,j'+2n)$
(recall that in Bayer pattern, downsampling by 2 in horizontal and vertical directions results in a single color image). Hence our overall goal is to learn optimal parameters using \eqref{eq:autoregression} and apply these parameters to the constrained estimator of \eqref{eq:estimate} and \eqref{eq:constraint}.

\subsection{Problem Formulation} \label{sec:ProbForm}

We formulate the problem of learning optimal weights $\vec{\alpha}=[\alpha_1,\dots,\alpha_L]^T\in\R^L$ in \eqref{eq:autoregression} in terms of image patches. Specifically, rewriting \eqref{eq:autoregression}, 
\begin{align}\label{eq:x'}
\widehat{x}' = (\vs_0-\vec{\beta}')^T \vec{C}\vec{\alpha}+\gamma
\end{align}
is an optimal estimate of $x_k(i',j')$, where $\vec{\beta},\vs_0\in\R^N$ are vectorized versions of $\sqrt{N}\times\sqrt{N}$ image patches taken from $\beta(m,n)$ and the downsampled image $s_k(i'+2m,j'+2n)=h(i'+2m,j'+2n)$ near $(i,j)$; $\gamma\in\R$ is $\gamma(0,0)$; and the column vectors of $\vec{C}\in\mathbb{R}^{N\times L}$ are the predetermined basis vectors in \eqref{eq:constraint}. In this work, we optimize $\vec{\alpha}$ in a total least squares (TLS) sense \cite{Golub96,Hirakawa06,Hirakawa06denoising} since TLS tolerates uncertainties in both an estimand $x$ and an observation $\vs_0$. 

\begin{figure}[!t]\begin{center}
\hspace{-1.0cm} \includegraphics[scale=0.3]{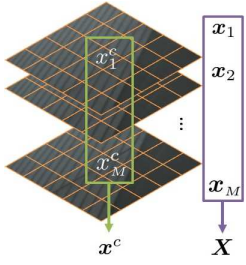}
\caption{The set of image patches collected from an image. \label{fig:patches}} 
\end{center} \end{figure}

Define the matrices $\mX \defequal \left[ \vx_1, \cdots, \vx_M \right] \in \R^{N \times M}$ and $\mS \defequal \left[ \vs_1, \cdots ,\vs_M\right] \in \R^{N \times M}$ as concatenations of $M$ vectorized $\sqrt{N}\times\sqrt{N}$ patches from the fog-free $\vx$ and foggy $\vs$ images, respectively. When these image patches $\{\vs_1,\dots,\vs_M\}$ are similar to the patch of interest $\vs_0$, we expect the relationship in \eqref{eq:x'} to hold:
\begin{align}
{\vx^c}^T\approx   (\mS-\vec{\beta}'\vec{1})^T\vec{C}\vec{\alpha} +\gamma\vec{1}^T,
\end{align}
where $\vec{1}=[1,\dots,1]\in\mathbb{R}^{1\times M}$, and $\vx^c\in\R^{1\times M}$ denotes the center pixels of the image patches in $\{\vx_1, \cdots, \vx_M\}$ (see Fig.~\ref{fig:patches}). When $M \gg N$, this system is algebraically overdetermined and it has no exact solution in general. Solving for $\vec{\alpha}$ in a TLS sense, we minimize the error $\vr\in\mathbb{R}^{1\times M}$ in estimating $\vx^c$ and error $\mE\in\mathbb{R}^{M\times N}$ in the measurement $\mS$ simultaneously in the following manner:
\begin{equation}\label{eq:GTLS}\begin{array}{l}
\left(\widehat{\mE},\widehat{\vr},\widehat{\vec{\beta}}',\widehat{\gamma}\right)=\underset{\mE,\vr,\vec{\beta}',\gamma}{\arg\min} \:\: \left\| \mA
\left[ \mE^T,\vr^T \right]
\begin{bmatrix}
\mC&\vec{0}\\
\vec{0}&1
\end{bmatrix}
 \mB \right\|_{\scriptscriptstyle{F}}^{2} \\\\
\mathrm{subject \: to} \:\:\: \vx^c-\gamma\vec{1}+\vr \in \calR\left( \mS-\vec{\beta}'\vec{1}+\mE \right),
\end{array}\end{equation}
where the symbol $\calR$ denotes the rowspace in $\mathbb{R}^M$ and $\|\cdot\|_{F}$ denotes the Frobenius matrix norm. Here, $\mA=\operatorname{diag}(a_1,\dots,a_M)\in\R^{M\times M}$ and $\mB=\operatorname{diag}(b_1,\dots,b_{L+1})\in\R^{(L+1)\times(L+1)}$ are predetermined weighting matrices. The triplet $(\widehat{\vec{\alpha}},\widehat{\beta}',\widehat{\gamma})$ satisfying
\begin{align}
{\vx^c}^T+\widehat{\vec{r}}^T =   (\mS-\widehat{\vec{\beta}}'\vec{1}+\widehat{\mE})^T\cdot \vec{C}\cdot\widehat{\vec{\alpha}} +\widehat{\gamma}\vec{1}^T
\end{align}
is known as the TLS solution. Once computed,  $(\widehat{\vec{\alpha}},\widehat{\beta}',\widehat{\gamma})$ can be used in \eqref{eq:x'} to yield a TLS-optimal estimation.

\subsection{TLS Solution} \label{sec:ProbSolving}

The solution to \eqref{eq:GTLS} is well-known when fog-free image $\vx$ is directly accessible~\cite{Golub96}. Define matrix $\mP$ and its singular decomposition as
\begin{align}\label{eq:P}
\begin{split}
\mP\defequal &\mA \left[(\mS-\vec{\beta}'\vec{1})^T,\: (\vx^{c}-\gamma\vec{1})^T \right] 
\begin{bmatrix}
\mC&\vec{0}\\
\vec{0}&1
\end{bmatrix}
\mB\\
=&\mU 
\Lambda
\mV^T,
\end{split}
\end{align}
where 
\begin{align} 
\vec{\beta}' &= \frac{1}{M}\sum_{m=1}^M \vs_m,\qquad
\gamma = \frac{1}{M}\sum_{m=1}^M x^c_m.
\end{align}
Then the TLS solution is given by:
\begin{align} \label{eq:SoltiontoGTLS}
\vec{\alpha} &= \frac{-1}{b_{\scriptscriptstyle{L+1}}v_{\scriptscriptstyle{L+1,L+1}}}\left[ \begin{array}{c} b_{\scriptscriptstyle 1}v_{\scriptscriptstyle{1,L+1}}\\ \vdots \\ b_{\scriptscriptstyle{L}}v_{\scriptscriptstyle{L,L+1}}\end{array}\right],
\end{align}
where $\vv_{L+1}$ is the right singular vector corresponding to the smallest singular value.

The TLS solution in \eqref{eq:SoltiontoGTLS} is impractical since the fog-free image $\vx$ is unobservable. The technique employed by the image denoising method in \cite{Hirakawa06denoising} and later adopted by the joint demosaicking and denoising method in \cite{Hirakawa06} is to introduce an auxiliary matrix $\mQ \in \R^{\left( N+1 \right) \times \left( N+1 \right)}$:
\begin{equation}\label{eq:Q}\begin{array}{ccl}
\mQ & = &\mP^T\mP \\ 
       & = & \left( \mU^{T} \Lambda \mV \right)^{T} \left( \mU^{T} \Lambda \mV \right) \vspace{0.2cm} \\
       & = & \mathbf{V}^{T} \Lambda^{2} \mathbf{V}\:.
\end{array} \end{equation} 
In other words, the right singular vector of $\mP$ can be interpreted as the eigenvectors of $\mQ$. The advantage of working with $\mQ$ over $\mP$ is that $\mQ$ can be estimated accurately. Substituting \eqref{eq:P} into \eqref{eq:Q}, $\mQ$ may be rewritten as
\begin{align} \label{eq:NewQ}
\mQ = \mB^{T}
\begin{bmatrix}
\mC^T&\vec{0}\\
\vec{0}&1
\end{bmatrix}
 \left[ \begin{array}{cc}
\mSigma_{\mS} & \mSigma_{\mS\vx^c}\\
 \mSigma_{\mS\vx^c}^{T} &  \mSigma_{\vx^c}
\end{array} \right] \begin{bmatrix}
\mC&\vec{0}\\
\vec{0}&1
\end{bmatrix}
\mB \:,
\end{align}
where the covariance matrices are defined by
\begin{align}\label{eq:sigma}
\begin{split}
\mSigma_{\mS}\defequal&(\mS-\vec{\beta}'\vec{1})\mA^2(\mS-\vec{\beta}'\vec{1})^T,\\
\mSigma_{\mS\vx^c}\defequal&(\mS-\vec{\beta}'\vec{1})\mA^2(\vx^c-\gamma\vec{1})^T,\\
\mSigma_{\vx^c}\defequal&(\vx^c-\gamma\vec{1})\mA^2(\vx^c-\gamma\vec{1}),
\end{split}
\end{align}
and $\mA^2 = \mA^{T} \mA$. 

Though $\mSigma_{\mS}$ can be computed directly from the observed data, $\mSigma_{\mS\vx^c}$ and $\mSigma_{\vx^c}$ must be obtained by indirect means. Invoking law of large numbers, we have the following relation:
\begin{align}\label{eq:xc}
\begin{array}{l}
\mSigma_{\mX} = \mathbb{E} \left[ (\mX-\vec{\beta}'\vec{1})\mA^2(\mX-\vec{\beta}'\vec{1})^T \right] \vspace{0.1cm} \\
\qquad = \mathbb{E} \left[ \operatorname{diag} (t)^{-1} (\mS-\vec{\beta}'\vec{1})\mA^2(\mS-\vec{\beta}'\vec{1})^T\operatorname{diag}(t)^{-1} \right.\\
\qquad\qquad \left. -\operatorname{diag}(\l_a+\sigma^2)\right], \vspace{0.2cm}\\
\mSigma_{\mS\mX}=\operatorname{diag} (t)\mSigma_{\mX}.
\end{array}
\end{align}
Taking the rows and columns corresponding to $\vx^c$ from $\mX$ yields $\mSigma_{\mS\vx^c}$ and $\mSigma_{\vx^c}$.

Following the steps to solve the problem of (\ref{eq:GTLS}), the estimate of $\widehat{x}'$ is obtained with~(\ref{eq:estimate}) and (\ref{eq:SoltiontoGTLS}). The overall method is summarized in Algorithm~\ref{alg1}.
\begin{algorithm}[h]\caption{Joint defogging and demosaicking algorithm} \label{alg1}
\begin{algorithmic}[1]
\STATE \textbf{Input:  $h$} 
\STATE \textbf{Output: $\widehat{x}'$}
\FOR{each pixel}
\STATE Define $\mS$ from a subsampled image $h$.
\STATE Compute $\mSigma_{\mS}$ by \eqref{eq:sigma}.
\STATE Estimate $\mSigma_{\mS\vx^c}$ and $\mSigma_{\vx^c}$ using \eref{eq:xc}. 
\STATE Compose $\mQ$ using \eqref{eq:Q}. 
\STATE Compute $\mV$ from eigen decomposition of $\mQ$.
\STATE Solve for $\vec{\alpha}$ in \eqref{eq:SoltiontoGTLS}.
\STATE Compute $\widehat{x}'$ in \eqref{eq:estimate}.
\ENDFOR
\end{algorithmic}
\end{algorithm}

\subsection{Implementation} \label{sec:Implementation}

\subsubsection{Weighting Matrices}
The weighting matrix $\mA$ controls perturbations of pixels in a patch and the matrix $\mB$ controls variations of a set of patches. Although the patches in a set of $\vs_m$ are chosen to have similar depth values, scene radiance may be dissimilar in a same depth layer. In order to measure the similarity of scene radiance between $\vs_0$ and $\vs_m$, the weighting matrix $\mA$ is defined as  
\begin{equation}
a_{m} = \exp \left( -\left( s_{0}^{c}-s_{m}^{c}\right)^{2}/ \kappa \right),
\end{equation}  
where $a_{m}$ is the $m$-th diagonal component of $\mA$. The subscript $c$ indicates the center values of patches and $\kappa$ is a constant. In the real sensor data experiments, we choose $b_{1}= \cdots = b_{L+1} =1$. \\

\subsubsection{Transmission Estimation}
\begin{figure}[!t] \begin{center}
\begin{minipage}{0.45\linewidth}
\centerline{\includegraphics[scale=0.30]{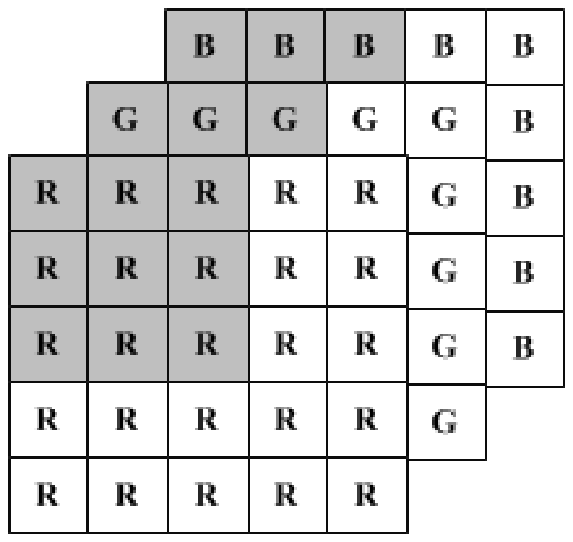}} 
\centerline{\footnotesize (a)}
\end{minipage}
\begin{minipage}{0.45\linewidth}
\centerline{\includegraphics[scale=0.33]{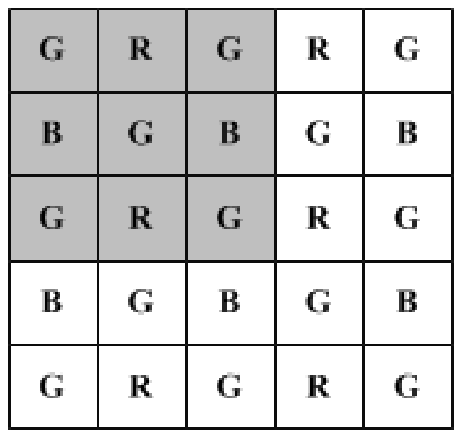}}
\centerline{\footnotesize (b)}
\end{minipage}
\caption{Measuring dark channels and minimum values among RGB channels in the shaded regions. (a) Region of measuring dark channels in~\eref{eq:dcp}. (b) Region of measuring dark channels in \eref{eq:estTransB}. \label{fig:DarkMinimum}}
\end{center}\end{figure}

In a fog-free patch, some color components are dark~\cite{He09}, known as {\it dark channel prior} :
\begin{align}\label{eq:dcp}
\min\limits_{\scriptscriptstyle{\vx(m,n) \in \Omega(i,j)}}\left( \min\limits_{\scriptscriptstyle{k \in \{ 1,2,3\}}} x_k(m,n)\right) \rightarrow 0,
\end{align}
where $\Omega(i,j)$ is a sample region centered at a pixel location $(i,j)$. Then transmission can be approximately estimated by taking the local minimum in a patch using \eref{eq:FoggyModel}: 
\begin{align}\label{eq:estTrans}
t(i,j) = 1-\rho \left( \min_{\scriptscriptstyle{\vy(m,n) \in \Omega(i,j)}} \left( \min_{\scriptscriptstyle{k \in \{ 1,2,3\}}} \frac{y_k(m,n)}{l_a}\right)\right),
\end{align} 
where the weighting coefficient $\rho$ controls the natural appearance of defogged images. Since not all color components are observable in practice, transmission in CFA sampled images are measured as followings (See \fref{fig:DarkMinimum}):
\begin{align}\label{eq:estTransB}
t(i,j) = 1-\rho \left( \min_{h(m,n) \in \Omega(i,j)} \frac{h(m,n)}{l_a}\right).
\end{align}
The {\it prior} distributions generated in RGB images and CFA sampled images are compared. The generated distributions demonstrate that the dark channel assumption is also valid in CFA images and transmission can be efficiently estimated in the newly defined region. In experiments throughout this paper, the window size is set to cover relatively large regions from $21 \times 21$ to $31 \times 31$~\cite{Jain88, Gibson13, Tarel09}; The weighting coefficient $\rho$ is set to $0.95$, typically used in conventional methods; Transmission is refined by dark pixel map to avoid halo effects in depth discontinuous regions~\cite{He13}. 

In order for our model to be effective, the set of patches $\vs_{k}$ are selected such that the features in $\vs_{0}$ are well preserved. The patches having similar transmission values are collected in the spatial vicinity of $\vs_{0}$. The similarity of transmission between patches is measured using the metric of Euclidean norm. Since pixels in collected patches may be located in different objects than that of a center pixel, we propose to prune outliers if the estimated values are unbounded~\cite{Hirakawa06}.

\section{Experimental Results}\label{sec:results}
We validate our model in Section~\ref{sec:algorithm} and the analysis and observations in Section~\ref{sec:background} on synthesized foggy images and raw image sensor data of real foggy scene. 

\subsection{Simulation Study}\label{sec:simStudy}
To verify theoretical analysis in Section~\ref{sec:background}, we created a test for noisy foggy images with synthesized depth, arbitrarily chosen scattering coefficients, and different noise levels. We first generated the foggy image $\vy(i,j)$ based on the model in \eqref{eq:FoggyModel}, where the scattering coefficients were in the range of $0.004(m^{-1})$ to $0.078(m^{-1})$ depending on the amount of fog~\cite{McCartney76}. The noise model in \eqref{eq:NoiseModel} was used to generate $\vs(i,j)$, the noisy version of $\vy(i,j)$. Here, $\sigma$ in \eref{eq:xc} was set to a very small value~(i.e. $l_a+\sigma^2 \approx l_a$). Finally, we synthesized a Bayer CFA-sampled data of foggy and noisy image using the model in \eqref{cfa2}.

The reconstruction of fog-free images by the proposed joint demosaicking and defogging method was implemented by taking $5 \times 5$ patches from a $25 \times 25$ neighborhood. The patch size was chosen as typical demosaicking filter size that generally gives the best performance among linear filters~\cite{Hirakawa06denoising,Malvar04}. Atmospheric light $\vl_a$ was estimated from the most haze-opaque regions using dark channels~\cite{He09}. For comparison, we applied well-known demosaicking (alternative projection~(AP)~\cite{Gunturk02, Lu10}) and defogging (dark channel prior~(DCP) \cite{He09} and Tarel's visibility restoration~\cite{Tarel09}) methods to the same input images. 

\begin{figure*}\begin{center}
\begin{tabular}{@{}>{\centering\arraybackslash}m{.12\linewidth}| >{\centering\arraybackslash}m{.12\linewidth}@{} @{}>{\centering\arraybackslash}m{.12\linewidth}@{} @{}>{\centering\arraybackslash}m{.12\linewidth}@{} @{}>{\centering\arraybackslash}m{.12\linewidth}@{} @{}>{\centering\arraybackslash}m{.12\linewidth}@{} @{}>{\centering\arraybackslash}m{.12\linewidth}@{} @{}>{\centering\arraybackslash}m{.12\linewidth}@{}}

\footnotesize Ground truth & \multicolumn{7}{c}{\footnotesize Thick fog}\\
\includegraphics[scale=0.5]{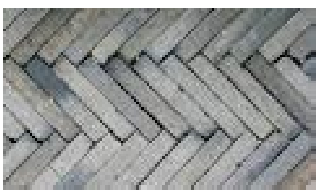} & \includegraphics[scale=0.5]{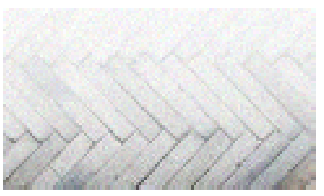} & \includegraphics[scale=0.5]{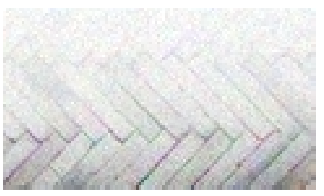} & \includegraphics[scale=0.5]{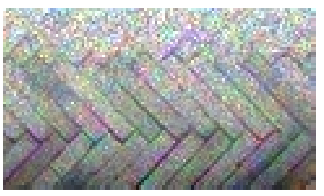} & \includegraphics[scale=0.5]{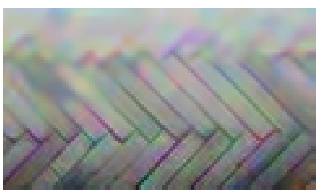} & \includegraphics[scale=0.5]{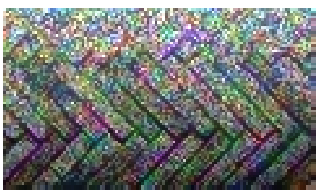} & \includegraphics[scale=0.5]{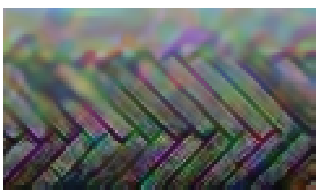} & \includegraphics[scale=0.5]{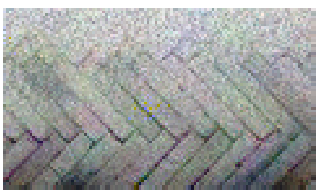} \\

\footnotesize Depth & \multicolumn{7}{c}{\footnotesize Moderate fog}\\
\includegraphics[scale=0.5]{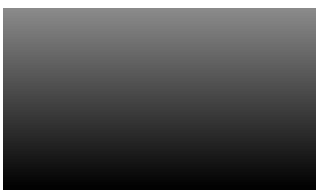} & \includegraphics[scale=0.5]{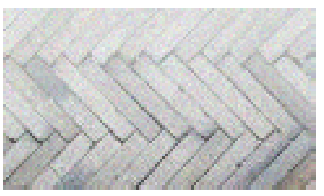} & \includegraphics[scale=0.5]{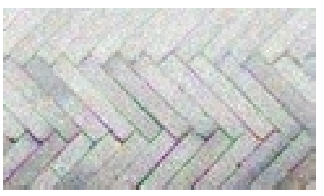} & \includegraphics[scale=0.5]{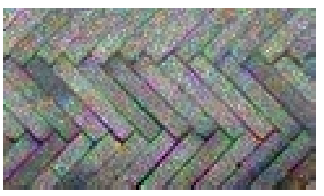} & \includegraphics[scale=0.5]{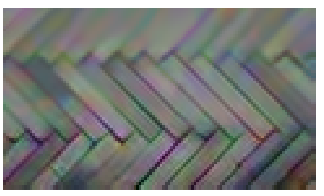} & \includegraphics[scale=0.5]{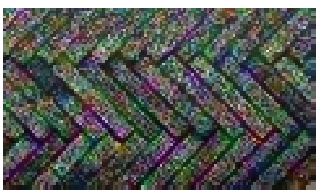} & \includegraphics[scale=0.5]{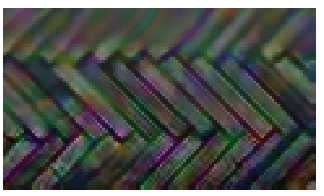} & \includegraphics[scale=0.5]{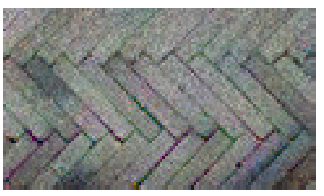}\\

& \multicolumn{7}{c}{\footnotesize Light fog}\\
& \includegraphics[scale=0.5]{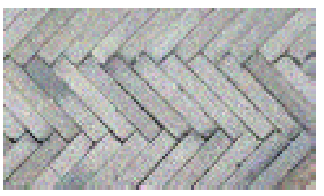} & \includegraphics[scale=0.5]{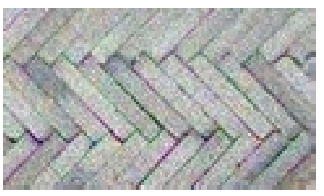} & \includegraphics[scale=0.5]{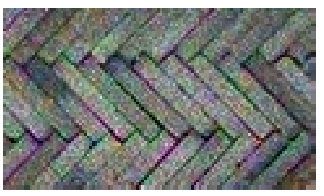} & \includegraphics[scale=0.5]{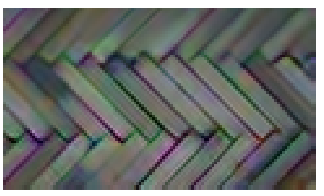} & \includegraphics[scale=0.5]{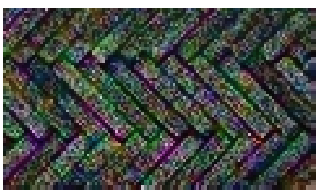} & \includegraphics[scale=0.5]{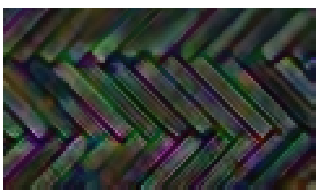} & \includegraphics[scale=0.5]{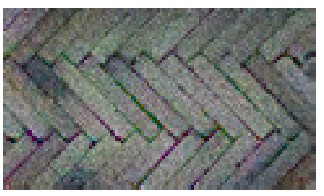}\\

& \footnotesize (a) & \footnotesize (b) & \footnotesize (c) & \footnotesize (d) & \footnotesize (e) & \footnotesize (f) & \footnotesize (g) \\
\end{tabular}
\caption{Examples of defogged images in a typical image processing pipeline with different level of shot noise variance and the readout noise, $\sigma = 0.01$. (a) Synthesized foggy and noisy images. (b) Input images demosaicked foggy and noisy images by AP~\cite{Gunturk02, Lu10}. (c) Defogged images by DCP method~\cite{He09}. (d) Output images denoised by block-matching and 3D filtering~(BM3D)~\cite{Dabov07} and defogged by DCP method. (e) Defogged images by Tarel's method~\cite{Tarel09}. (f) Output images denoised by block-matching and 3D filtering~(BM3D)~\cite{Dabov07} and defogged by Tarel's method. (g) Defogged images by the proposed method. In this simulation, $b_{L+1}$ was set to $0.5$ allowing larger perturbations between collected patches so that the choice of an optimal $\alpha$ is less influenced by their dissimilarities. \label{fig:DefoggedWithAlphas}}
\end{center}\end{figure*}

The simulation results as shown in Fig.~\ref{fig:DefoggedWithAlphas} validate our theoretical analysis which implies that the influence of noise is greatly amplified in distance scenes of the restored images. In the examples, noise clearly affects the restored images although it is not noticeable in the input images. Taking into account noise amplification in distant scenes, shutter speed should increase in order to prevent brightest pixels from saturating as fog becomes denser; a faster shutter produces more noise in the output image of a sensor. The examples in \fref{fig:compDenoising} clearly show that a typical image processing pipeline may fail at distant scene. In addition, denoising process degrades visual quality of the restored images. 

\begin{figure}[!t]\begin{center}
\begin{minipage}{0.49\linewidth}
\centerline{\includegraphics[scale=0.24]{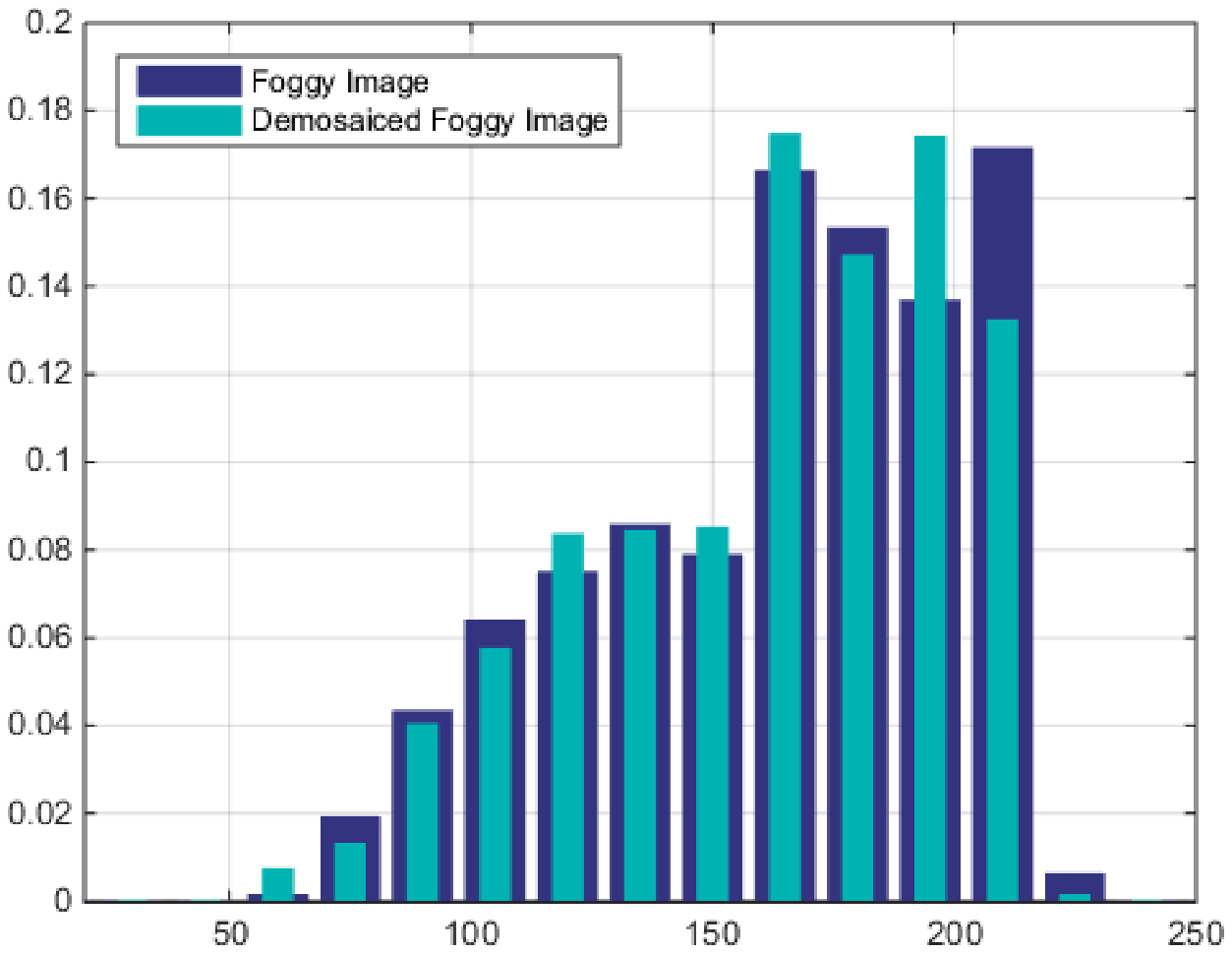}}
\centerline{\footnotesize (a)}
\end{minipage}
\begin{minipage}{0.49\linewidth}
\centerline{\includegraphics[scale=0.24]{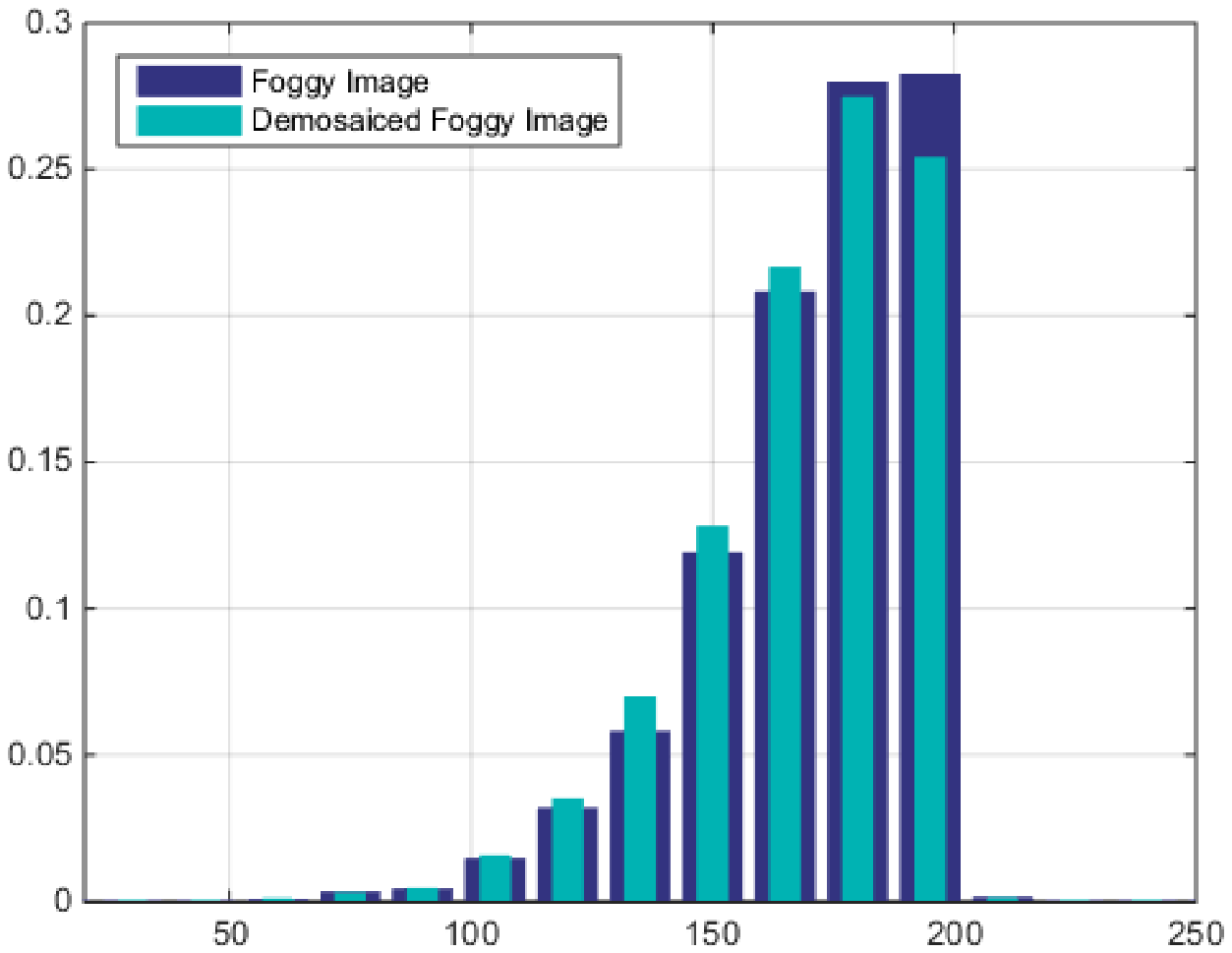}}
\centerline{\footnotesize (b)}
\end{minipage}
\caption{An example of statistics changes of foggy images and demosaicked foggy images. (a) Changes on statistics of Dark channels in DCP method. (b) Changes on atmospheric veiling in Tarel's method. \label{fig:radiance_change_demosacking}}
\end{center} \end{figure}

Similar to sensor noise amplification by the defogging process, demosaicking artifacts are magnified and clearly visible after applying a defogging algorithm. Moreover, defogging performance varies with the change of spatial statistics of the demosaicked foggy images. The small change of physical characteristics of foggy images by demosaicking~(see Fig.~\ref{fig:radiance_change_demosacking}) can result in large differences in estimating transmission, and this eventually increases uncertainty in the estimated radiance. These synergistic effects of defogging and demosaicking imply that processing them jointly is more effective than treating them separately. 

\subsection{Real Sensor Results}\label{sec:sensorResults}
We repeated the experiment by directly processing raw sensor data capturing a real foggy scene. We captured the foggy scenes using PointGrey Grasshopper3 in raw mode. The acquired raw data were processed following a typical digital camera processing pipeline in Fig.~\ref{fig:DSC}. The equalization/white balance was implemented using gray world~\cite{Land83}. The gamma correction was modest~($\gamma = 1.25$). 

\begin{figure*}[!t] \begin{center}
\begin{minipage}{0.19\linewidth}
\centerline{\includegraphics[scale=1.0]{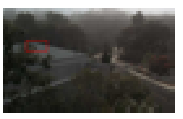}}
\centerline{\footnotesize{}} \vspace{0.3cm}
\end{minipage}
\begin{minipage}{0.19\linewidth}
\centerline{\includegraphics[scale=1.0]{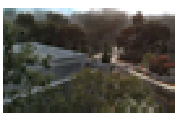}}
\centerline{\footnotesize{$(0.21, 2.31)$}} \vspace{0.2cm}
\end{minipage}
\begin{minipage}{0.19\linewidth}
\centerline{\includegraphics[scale=1.0]{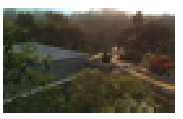}}
\centerline{\footnotesize{$(0.56, 2.60)$}} \vspace{0.2cm}
\end{minipage}
\begin{minipage}{0.19\linewidth}
\centerline{\includegraphics[scale=1.0]{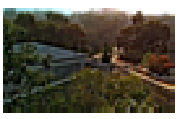}}
\centerline{\footnotesize{$(0.10, 3.59)$}} \vspace{0.2cm}
\end{minipage}
\begin{minipage}{0.19\linewidth}
\centerline{\includegraphics[scale=1.0]{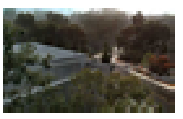}}
\centerline{\footnotesize{$(0.20, 2.74)$}} \vspace{0.2cm}
\end{minipage}
\\
\begin{minipage}{0.19\linewidth}
\centerline{\includegraphics[scale=1.0]{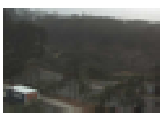}}
\centerline{\footnotesize{}}
\centerline{\footnotesize{(a)}}
\end{minipage}
\begin{minipage}{0.19\linewidth}
\centerline{\includegraphics[scale=1.0]{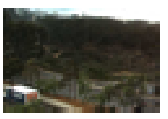}}
\centerline{\footnotesize{$(1.47, 2.41)$}}
\centerline{\footnotesize{(b)}}
\end{minipage}
\begin{minipage}{0.19\linewidth}
\centerline{\includegraphics[scale=1.0]{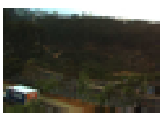}}
\centerline{\footnotesize{$(1.94,2.38)$}}
\centerline{\footnotesize{(c)}}
\end{minipage}
\begin{minipage}{0.19\linewidth}
\centerline{\includegraphics[scale=1.0]{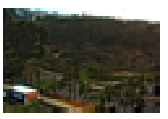}}
\centerline{\footnotesize{$(1.39,3.21)$ }}
\centerline{\footnotesize{(d)}}
\end{minipage}
\begin{minipage}{0.19\linewidth}
\centerline{\includegraphics[scale=1.0]{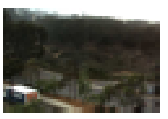}}
\centerline{\footnotesize{$(1.58,2.24)$}}
\centerline{\footnotesize{(e)}}
\end{minipage}
\caption{Examples of joint defogging and demosaicking. The values of below images represent the blind contrast measurement scores~$(e, \bar{r})$. (a) White-balanced input images~(Demosaicked by LSLCD method). (b) Defogged images by DCP method. (c) Defogged images by Tarel's method. (d) Defogged images by Wiener filter method. (e) Defogged images by the proposed TLS method. \label{fig:ResultsDemosaickingDefogging}}
\end{center}\end{figure*} 

\begin{figure*} \begin{center}
\begin{tabular}{m{0.2cm} >{\centering\arraybackslash}m{.14\textwidth} @{}>{\centering\arraybackslash}m{.14\textwidth} @{}>{\centering\arraybackslash}m{.14\textwidth} @{}>{\centering\arraybackslash}m{.14\textwidth} @{}>{\centering\arraybackslash}m{.14\textwidth}| >{\centering\arraybackslash}m{.14\textwidth}}

& \footnotesize{AHD~\cite{Hirakawa05ahd}} & \footnotesize{DLMMSE~\cite{Zhang05}} &\footnotesize{PDF~\cite{Menon07}} & \footnotesize{AP~\cite{Gunturk02, Lu10}} & \footnotesize{LSLCD~\cite{Dubois11}} & \footnotesize{Proposed} \\

\footnotesize{(a)}& \includegraphics[scale=0.8]{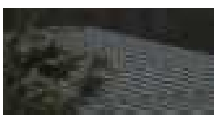} & \includegraphics[scale=0.8]{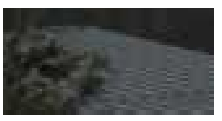} & \includegraphics[scale=0.8]{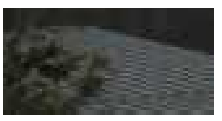} & \includegraphics[scale=0.8]{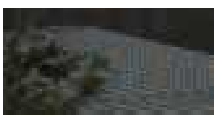} & \includegraphics[scale=0.8]{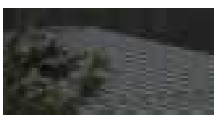} & \\ 

\footnotesize{(b)}& \includegraphics[scale=0.8]{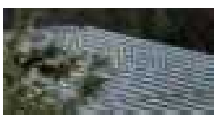} & \includegraphics[scale=0.8]{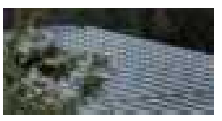} & \includegraphics[scale=0.8]{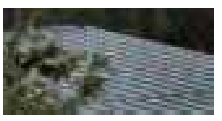} & \includegraphics[scale=0.8]{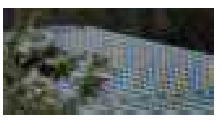} & \includegraphics[scale=0.8]{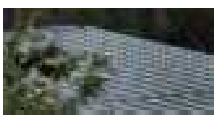} & \includegraphics[scale=0.8]{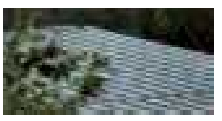} \\ 

\footnotesize{(c)}& \includegraphics[scale=0.8]{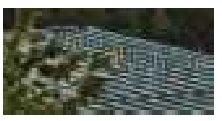} & \includegraphics[scale=0.8]{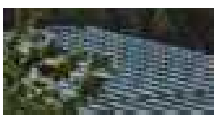} & \includegraphics[scale=0.8]{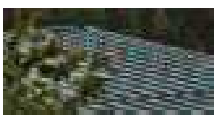} & \includegraphics[scale=0.8]{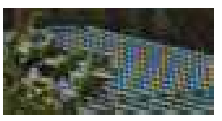} & \includegraphics[scale=0.8]{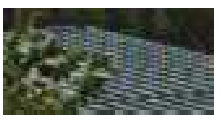} & \\ 

\footnotesize{(d)}& \includegraphics[scale=0.8]{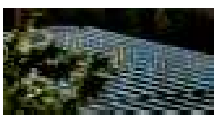} & \includegraphics[scale=0.8]{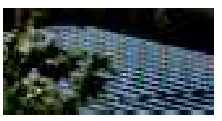} & \includegraphics[scale=0.8]{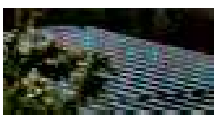} & \includegraphics[scale=0.8]{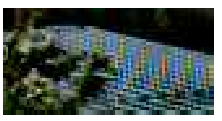} & \includegraphics[scale=0.8]{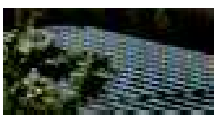} & \\ 

\footnotesize{(e)}& \includegraphics[scale=0.8]{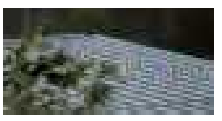} & \includegraphics[scale=0.8]{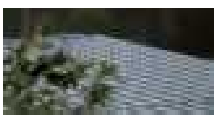} & \includegraphics[scale=0.8]{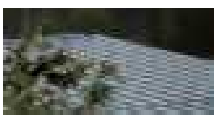} & \includegraphics[scale=0.8]{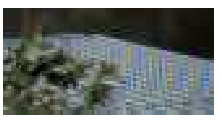} & \includegraphics[scale=0.8]{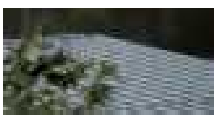} & \\

\footnotesize{(f)}& \includegraphics[scale=0.8]{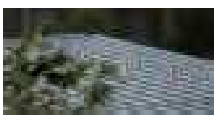} & \includegraphics[scale=0.8]{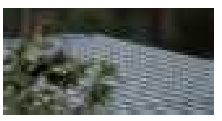} & \includegraphics[scale=0.8]{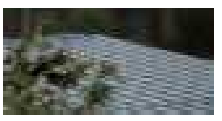} & \includegraphics[scale=0.8]{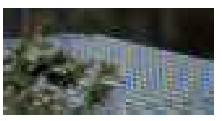} & \includegraphics[scale=0.8]{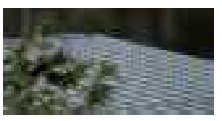} & \\

\footnotesize{(g)}& \includegraphics[scale=0.8]{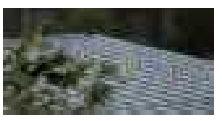} & \includegraphics[scale=0.8]{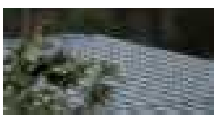} & \includegraphics[scale=0.8]{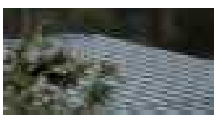} & \includegraphics[scale=0.8]{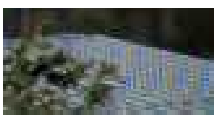} & \includegraphics[scale=0.8]{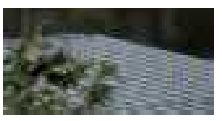} & \\

\end{tabular}
\caption{Visual comparison of various defogging, demosaicking, and denoising algorithms in the region of interest~(red box) in~\fref{fig:ResultsDemosaickingDefogging}. The output image of the proposed method is shown on the right side. (a) Demosaicked foggy images. (b) Defogged~(DCP) images. (c) Defogged~(Tarel's) images. (d) Defogged~(Wiener filter) images. (e) Denoised~(NLM~\cite{Buades05}) and defogged~(DCP) images. (f) Denoised~(BM3D~\cite{Dabov07}) and defogged~(DCP) images. (g) Denoised~(EPLL~\cite{Zoran11}) and defogged~(DCP) images. \label{fig:jointROI}}
\end{center} \end{figure*}  

\begin{figure*}[!t]\begin{center}
\begin{tabular}{@{}>{\centering\arraybackslash}m{.02\textwidth} 
>{\centering\arraybackslash}m{.24\textwidth}@{} 
@{}>{\centering\arraybackslash}m{.24\textwidth}@{} 
@{}>{\centering\arraybackslash}m{.24\textwidth}@{}
@{}>{\centering\arraybackslash}m{.24\textwidth}@{}}

& \footnotesize Foggy~(LSLCD) & \footnotesize DCP~\cite{He09} & \footnotesize Tarel's~\cite{Tarel09} & \footnotesize  Wiener~\cite{Gibson13} \\ 

\footnotesize (a) & \includegraphics[scale=1.0]{foggy-lslcd-rectangle} & \includegraphics[scale=1.0]{dcp-lslcd} & \includegraphics[scale=1.0]{tarel-lslcd} & \includegraphics[scale=1.0]{wiener-lslcd} \\ \hline \\ 

\footnotesize (b) & \includegraphics[scale=1.0]{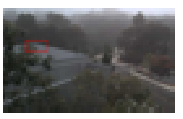} & \includegraphics[scale=1.0]{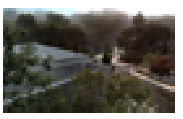} & \includegraphics[scale=1.0]{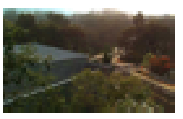} &  \includegraphics[scale=1.0]{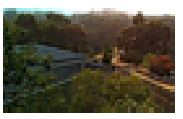} \\
& & \footnotesize (0.28, 2.01) & \footnotesize (1.13, 2.37) & \footnotesize (0.36, 3.03)\\

\footnotesize (c) & \includegraphics[scale=1.0]{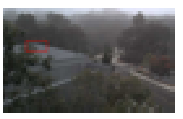} & \includegraphics[scale=1.0]{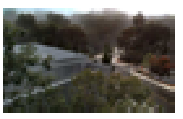} & \includegraphics[scale=1.0]{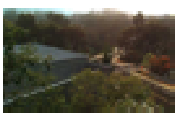} & \includegraphics[scale=1.0]{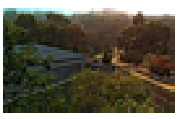} \\ 
& & \footnotesize (0.42, 1.98) & \footnotesize (0.62, 2.33) & \footnotesize (0.41, 3.26) \\

\footnotesize (d) & \includegraphics[scale=1.0]{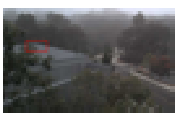} & \includegraphics[scale=1.0]{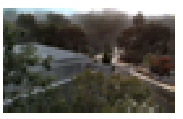} & \includegraphics[scale=1.0]{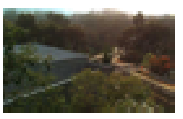} & \includegraphics[scale=1.0]{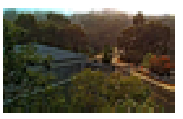} \\
& & \footnotesize (0.43, 1.93) & \footnotesize (0.76, 2.33) & \footnotesize (0.47, 3.29)\\ \hline \\

\footnotesize (e) & \includegraphics[scale=1.0]{foggy-lslcd-rectangle} & \includegraphics[scale=1.0]{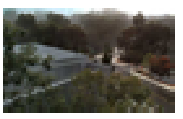} & \begin{align*}\begin{array}{c}\footnotesize \textrm{NLM:}\:(0.85,\:2.52) \\ \footnotesize \textrm{BM3D:}\:(0.67,\:2.21) \\ \footnotesize \textrm{EPLL:}\: (0.57,\:2.04) \end{array}\end{align*} & \\
\end{tabular}
\caption{Comparison of joint defogging and demosaicking with different demosaicking and denoising algorithms. The values of below images represent the blind contrast measurement scores $e$ and $\bar{r}$. All foggy images are demosaicked by LSLCD method. The foggy images in the {\it a}-th and {\it e}-th rows are demosaicked without applying denoising. The scores of the proposed method are measured using the foggy images in the {\it b}-th row to the {\it d}-th row. (a) Defogged images without denoising. (b) Denoised~(NLM) and defogged images. (c) Denoised~(BM3D) and defogged images. (d) Denoised~(EPLL) and defogged images. (e) Defogged image by the proposed method. \label{fig:compDenoising}}
\end{center}\end{figure*}

Working in the raw sensor domain ensures that the fog model in \eqref{eq:FoggyModel} and the noise model in \eqref{eq:NoiseModel} hold, since sensor data are approximately linear with respect to light. Demosaicking is also designed to work in pre-gamma correction domain. To the best of our knowledge, this is the first study of combining defogging and demosaicking algorithms. Therefore, the proposed joint algorithm was compared against the state-of-the-art defogging algorithms for the acquired raw sensor data. Among these are: DCP method~\cite{He09}, Tarel's method~\cite{Tarel09}, and Wiener filter approach~\cite{Gibson13}. To implement a typical pipeline, the acquired raw sensor data were reconstructed by following demosaicking algorithms: adaptive homogeneity-directed demosaicking~(AHD)~\cite{Hirakawa05ahd}, directional linear minimum mean square-error estimation~(DLMMSE)~\cite{Zhang05}, posterior directional filtering~(PDF)~\cite{Menon07}, AP~\cite{Gunturk02, Lu10}, and LSLCD~\cite{Dubois11}. Then, the comparing defogging algorithms were applied to the demosaicked images. For both the results of the typical pipeline and the proposed algorithm, gamma correction was applied after recovering foggy images for display. The blind contrast measurements~\cite{Hautiere08} were used to show quantitative evaluation as well as perceptual qualitative comparison. The metric $e$ represents the restored visible edges that were invisible in foggy images. The value of $\bar{r}$ is the ratio of the gradients norms before and after restoration at visible edges. The visible edges were selected by a $5\%$ contrast threshold in daytime fog. It is important to note that further color adjustment or tone reproduction algorithms were not applied to the defogged images in order to thoroughly study defogging performance without any secondary effects. The visual quality of the resulting images can be further adjusted to the human eye if other image enhancement algorithms are applied. 

\begin{figure*}[!t] \begin{center}
\begin{minipage}{0.19\linewidth}
\centerline{\includegraphics[scale=1.0]{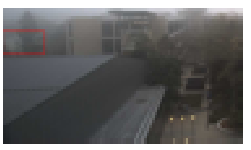}}
\vspace{0.1cm}
\end{minipage}
\begin{minipage}{0.19\linewidth}
\centerline{\includegraphics[scale=1.0]{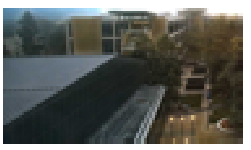}}
\vspace{0.1cm}
\end{minipage}
\begin{minipage}{0.19\linewidth}
\centerline{\includegraphics[scale=1.0]{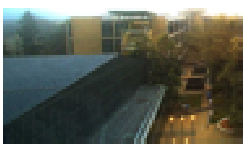}}
\vspace{0.1cm}
\end{minipage}
\begin{minipage}{0.19\linewidth}
\centerline{\includegraphics[scale=1.0]{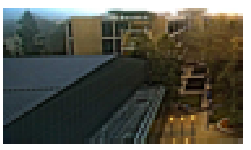}}
\vspace{0.1cm}
\end{minipage}
\begin{minipage}{0.19\linewidth}
\centerline{\includegraphics[scale=1.0]{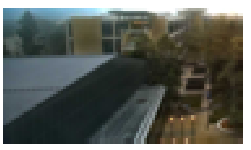}}
\vspace{0.1cm}
\end{minipage}
\\
\begin{minipage}{0.19\linewidth}
\centerline{\includegraphics[scale=0.3]{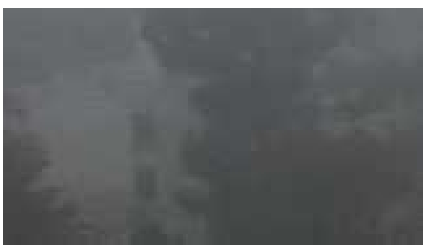} \includegraphics[scale=0.3]{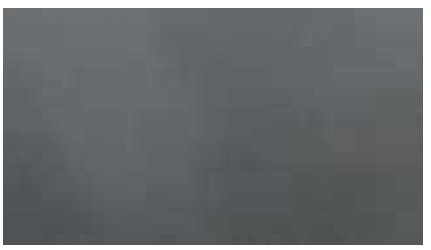}}
\vspace{0.1cm}
\end{minipage}
\begin{minipage}{0.19\linewidth}
\centerline{\includegraphics[scale=0.3]{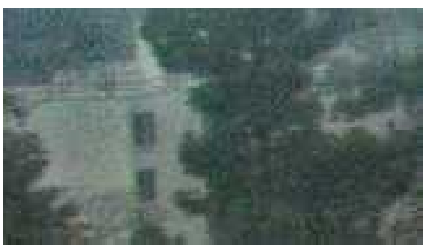} \includegraphics[scale=0.3]{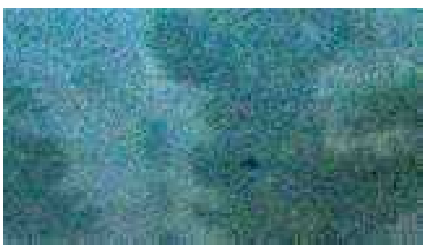}}
\vspace{0.1cm}
\end{minipage}
\begin{minipage}{0.19\linewidth}
\centerline{\includegraphics[scale=0.3]{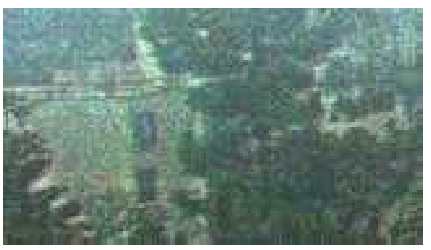} \includegraphics[scale=0.3]{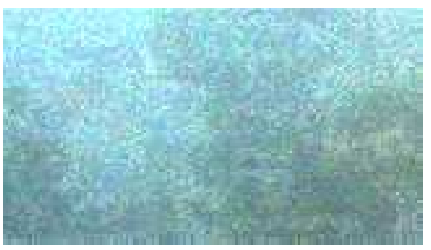}}
\vspace{0.1cm}
\end{minipage}
\begin{minipage}{0.19\linewidth}
\centerline{\includegraphics[scale=0.3]{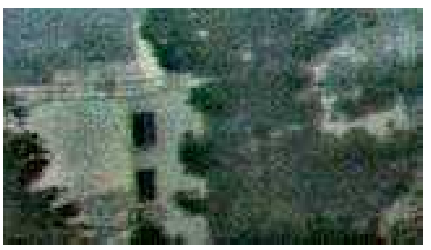} \includegraphics[scale=0.3]{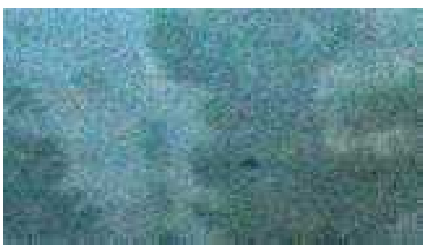}}
\vspace{0.1cm}
\end{minipage}
\begin{minipage}{0.19\linewidth}
\centerline{\includegraphics[scale=0.3]{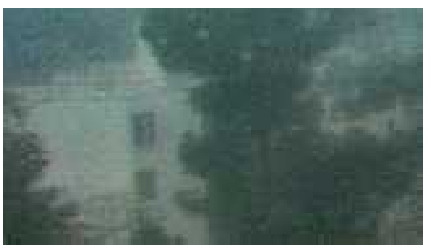} \includegraphics[scale=0.3]{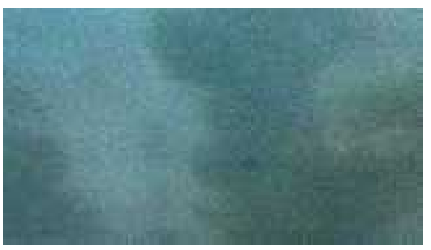}}
\vspace{0.1cm}
\end{minipage}
\\
\begin{minipage}{0.19\linewidth}
\centerline{\includegraphics[scale=1.0]{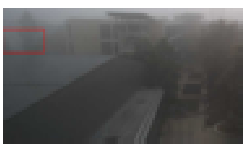}}
\centerline{\footnotesize{(a)}}
\end{minipage}
\begin{minipage}{0.19\linewidth}
\centerline{\includegraphics[scale=1.0]{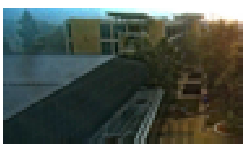}}
\centerline{\footnotesize{(b)}}
\end{minipage}
\begin{minipage}{0.19\linewidth}
\centerline{\includegraphics[scale=1.0]{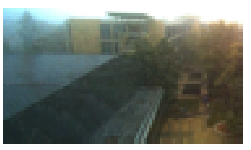}}
\centerline{\footnotesize{(c)}}
\end{minipage}
\begin{minipage}{0.19\linewidth}
\centerline{\includegraphics[scale=1.0]{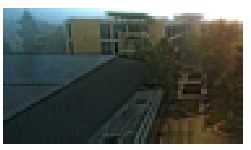}}
\centerline{\footnotesize{(d)}}
\end{minipage}
\begin{minipage}{0.19\linewidth}
\centerline{\includegraphics[scale=1.0]{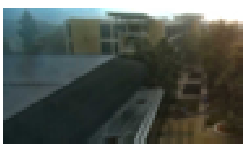}}
\centerline{\footnotesize{(e)}}
\end{minipage}
\caption{Examples of joint defogging and demosaicking in different foggy conditions. The images in the middle row are the results of the regions of interests. The images on the left side of each column are cropped from top images and the right images are from the bottom images. We increased exposure of the zoomed images. (a) White-balanced input images demosaicked by DLMMSE method. (b) Defogged images by DCP method. (c) Defogged images by Tarel's method. (d) Defogged images by Wiener filter method. (e) Defogged image by proposed TLS method.\label{fig:ResultsDemosaickingDefogging2}}
\end{center} \end{figure*}

\begin{figure*}\begin{center}
\begin{tabular}{@{}>{\centering\arraybackslash}m{.02\linewidth}@{\:} 
@{}>{\centering\arraybackslash}m{.097\linewidth}@{} @{\:}>{\centering\arraybackslash}m{.097\linewidth}@{} 
@{}>{\centering\arraybackslash}m{.097\linewidth}@{} @{\:}>{\centering\arraybackslash}m{.097\linewidth}@{}
@{}>{\centering\arraybackslash}m{.097\linewidth}@{} @{\:}>{\centering\arraybackslash}m{.097\linewidth}@{}
@{}>{\centering\arraybackslash}m{.097\linewidth}@{} @{\:}>{\centering\arraybackslash}m{.097\linewidth}|
>{\centering\arraybackslash}m{.097\linewidth}@{} @{\:}>{\centering\arraybackslash}m{.097\linewidth}@{}}

& \multicolumn{2}{@{}c@{}}{\includegraphics[scale=1.0]{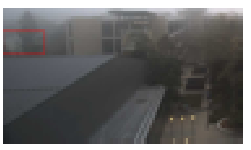}} & \multicolumn{2}{@{}c@{}}{\includegraphics[scale=1.0]{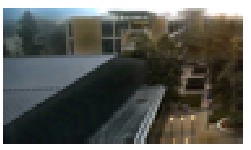}} & \multicolumn{2}{@{}c@{}}{\includegraphics[scale=1.0]{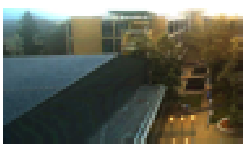}} & \multicolumn{2}{@{}c|}{\includegraphics[scale=1.0]{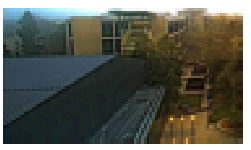}}& \multicolumn{2}{c@{}}{}\\

\footnotesize (a) & \includegraphics[scale=0.295]{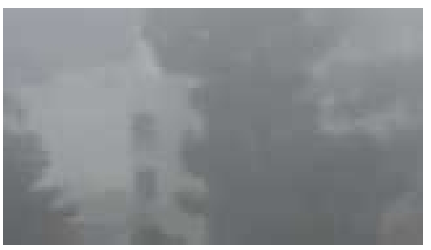} & \includegraphics[scale=0.295]{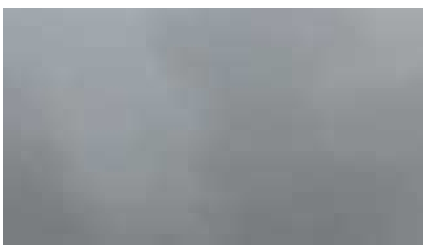} & \includegraphics[scale=0.295]{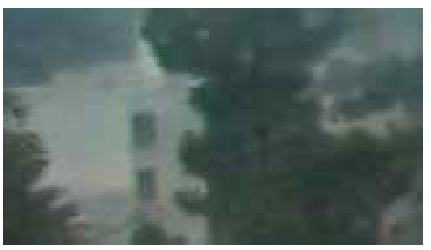} 
& \includegraphics[scale=0.295]{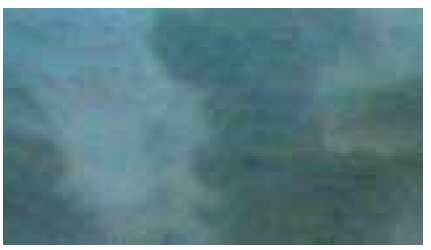} & \includegraphics[scale=0.295]{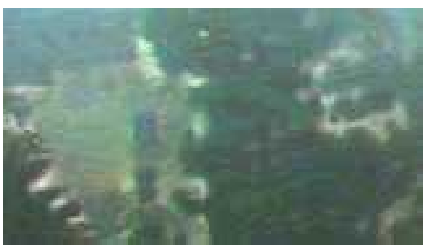} & \includegraphics[scale=0.295]{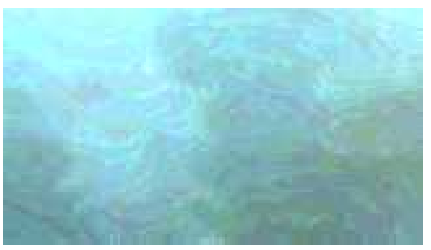} & \includegraphics[scale=0.295]{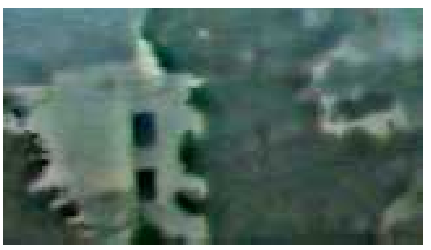} & \includegraphics[scale=0.295]{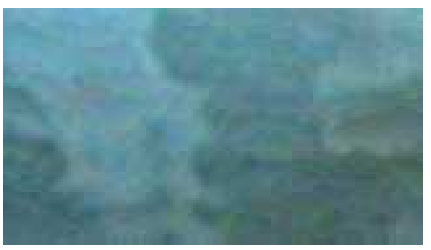} & & \\

& \multicolumn{2}{@{}c@{}}{\includegraphics[scale=1.0]{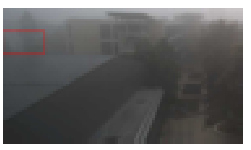}} & \multicolumn{2}{@{}c@{}}{\includegraphics[scale=1.0]{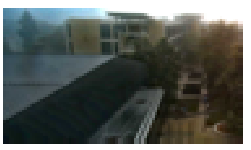}} & \multicolumn{2}{@{}c@{}}{\includegraphics[scale=1.0]{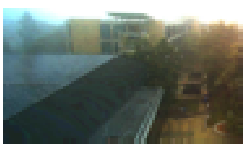}} & \multicolumn{2}{@{}c|}{\includegraphics[scale=1.0]{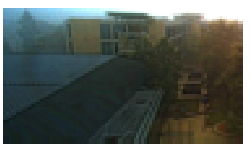}} & \multicolumn{2}{c@{}}{}\\ 

& & & & & & & & & & \\ \cline{1-9}
& & & & & & & & & & \\   

& \multicolumn{2}{@{}c@{}}{\includegraphics[scale=1.0]{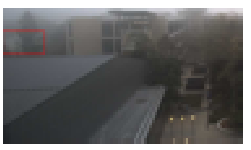}} & \multicolumn{2}{@{}c@{}}{\includegraphics[scale=1.0]{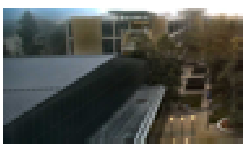}} & \multicolumn{2}{@{}c@{}}{\includegraphics[scale=1.0]{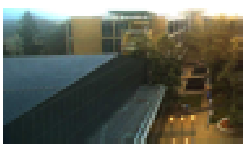}} & \multicolumn{2}{@{}c|}{\includegraphics[scale=1.0]{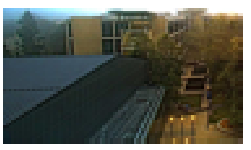}}& \multicolumn{2}{c@{}}{\includegraphics[scale=1.0]{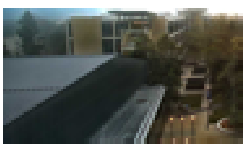}}\\

\footnotesize (b) & \includegraphics[scale=0.285]{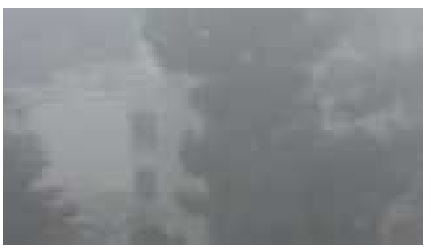} & \includegraphics[scale=0.295]{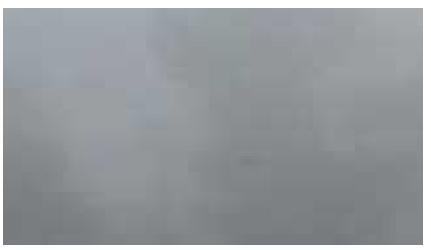} & \includegraphics[scale=0.295]{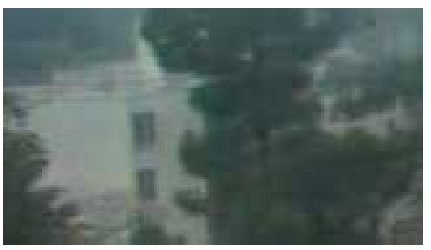} 
& \includegraphics[scale=0.295]{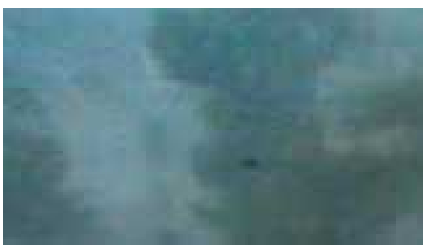} & \includegraphics[scale=0.295]{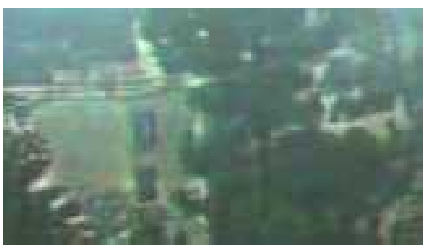} & \includegraphics[scale=0.295]{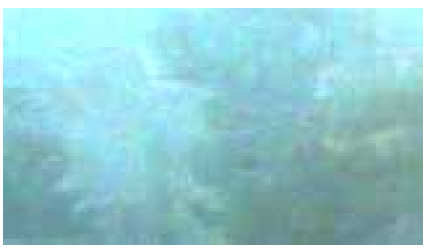} & \includegraphics[scale=0.295]{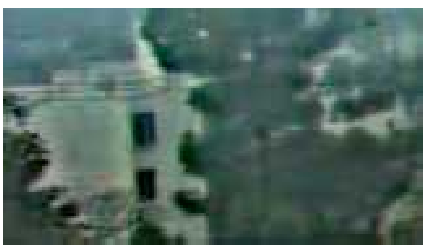} & \includegraphics[scale=0.295]{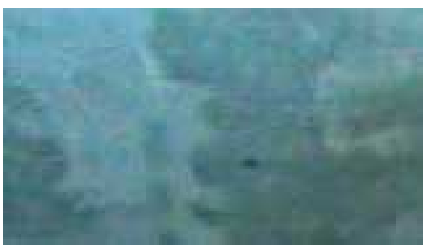} & \includegraphics[scale=0.295]{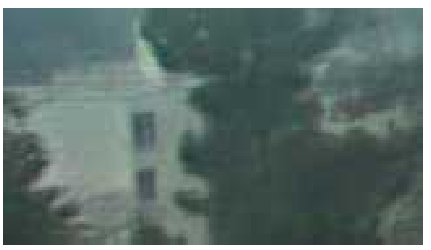} & \includegraphics[scale=0.295]{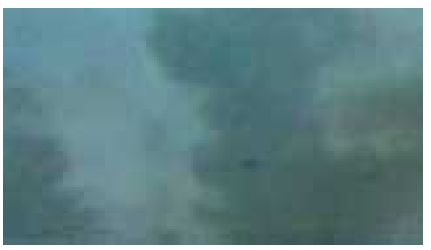} \\

& \multicolumn{2}{@{}c@{}}{\includegraphics[scale=1.0]{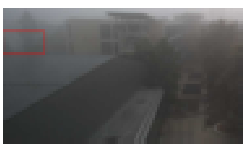}} & \multicolumn{2}{@{}c@{}}{\includegraphics[scale=1.0]{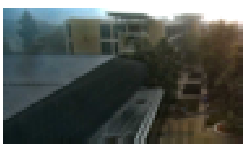}} & \multicolumn{2}{@{}c@{}}{\includegraphics[scale=1.0]{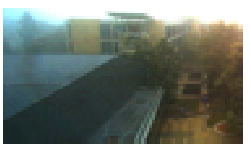}} & \multicolumn{2}{@{}c|}{\includegraphics[scale=1.0]{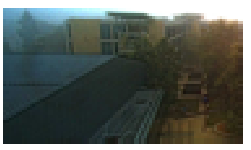}} & \multicolumn{2}{c@{}}{\includegraphics[scale=1.0]{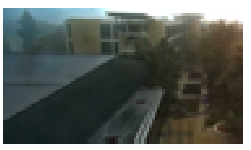}}\\ 

& & & & & & & & & & \\ \cline{1-9}
& & & & & & & & & & \\ 

& \multicolumn{2}{@{}c@{}}{\includegraphics[scale=1.0]{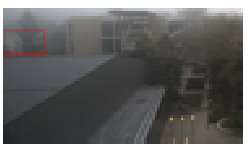}} & \multicolumn{2}{@{}c@{}}{\includegraphics[scale=1.0]{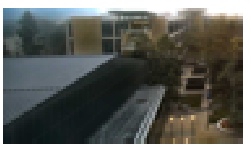}} & \multicolumn{2}{@{}c@{}}{\includegraphics[scale=1.0]{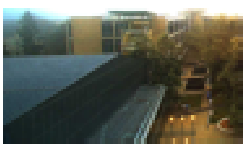}} & \multicolumn{2}{@{}c|}{\includegraphics[scale=1.0]{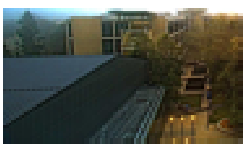}}& \multicolumn{2}{c@{}}{}\\

\footnotesize (c) & \includegraphics[scale=0.295]{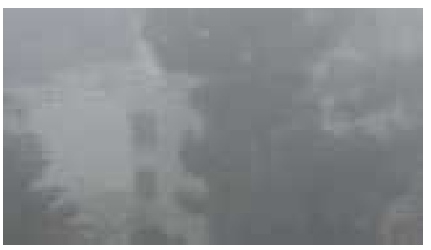} & \includegraphics[scale=0.295]{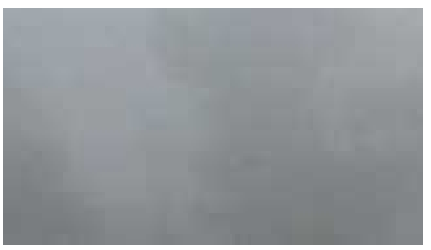} & \includegraphics[scale=0.295]{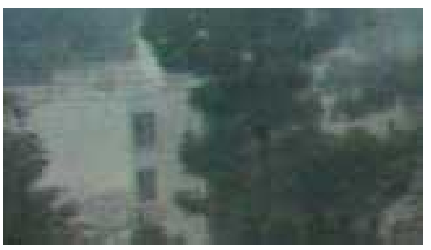} 
& \includegraphics[scale=0.295]{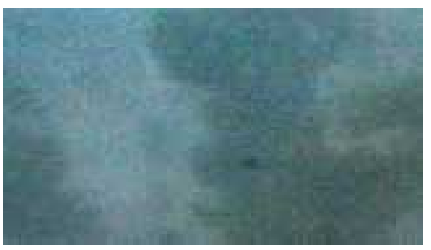} & \includegraphics[scale=0.295]{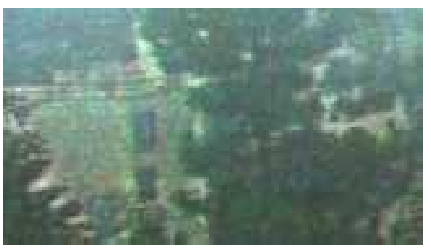} & \includegraphics[scale=0.295]{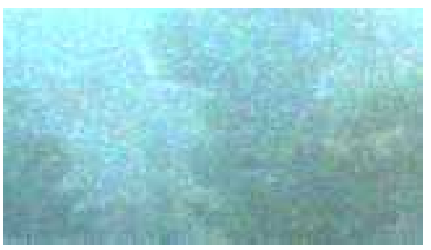} & \includegraphics[scale=0.295]{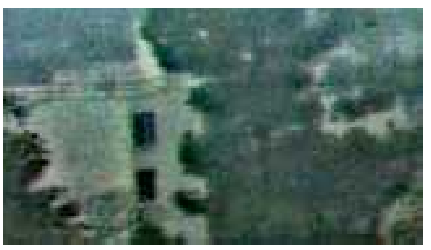} & \includegraphics[scale=0.295]{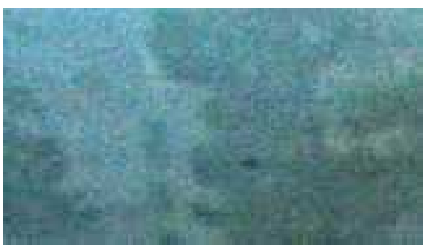} & & \\

& \multicolumn{2}{@{}c@{}}{\includegraphics[scale=1.0]{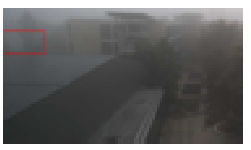}} & \multicolumn{2}{@{}c@{}}{\includegraphics[scale=1.0]{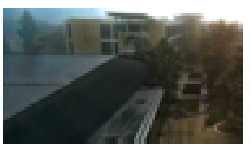}} & \multicolumn{2}{@{}c@{}}{\includegraphics[scale=1.0]{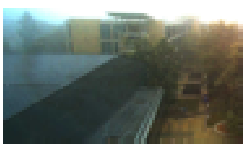}} & \multicolumn{2}{@{}c|}{\includegraphics[scale=1.0]{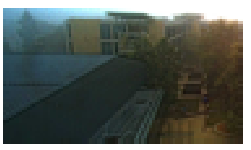}} & \multicolumn{2}{c@{}}{}\\ [1ex]

& \multicolumn{2}{@{}c@{}}{\footnotesize Foggy} & \multicolumn{2}{@{}c@{}}{\footnotesize DCP} & \multicolumn{2}{@{}c@{}}{\footnotesize Tarel's} &\multicolumn{2}{@{}c|}{\footnotesize Wiener filter} & \multicolumn{2}{c@{}}{\footnotesize Proposed method}\\[2ex]

\end{tabular}
\caption{Comparisons of joint defogging and demosaicking with denoised outputs in different foggy conditions. The comparing results were processed by the processing order of a typical pipeline--demosaicking, denoising, and defogging. The input foggy images are demosaicked by DLMMSE method. (a) Denoised output images by NLM. (b) Denoised output images by BM3D. (c) Denoised output images by EPLL. \label{fig:Denoising2}}
\end{center}\end{figure*}

The examples of output images of demosaicking and defogging are shown in \fref{fig:ResultsDemosaickingDefogging} and Fig.~\ref{fig:ResultsDemosaickingDefogging2}. Even though the blind contrast measurement scores are higher in the output images of Tarel's method or Wiener filter approach, they are extremely oversaturated and their scene colors are unnaturally distorted. This implies that those methods stretch local contrast and produce spurious edges. Though the blind contrast measurements show improvement on gradients, a higher value does not guarantee better perceptual quality~\cite{Ancuti13}. The most noticeable demosaicking artifact occurs on the gray roof of the red box in \fref{fig:ResultsDemosaickingDefogging}. The cropped regions of the results are displayed in Fig.~\ref{fig:jointROI}. Zippering artifacts become apparent after applying defogging algorithms and more visible in the output images restored by Tarel's and Wiener filter methods. The examples clearly demonstrate that demosaicking artifacts become worse when treating defogging and demosaicking separately than combining them. The second test image in \fref{fig:ResultsDemosaickingDefogging} is a difficult task for the proposed algorithm, as a bright object and a dark object are overlayed in a small region around the trail. The effectiveness of the proposed algorithm depends on the similarity of collected patches; thus, patches should be carefully selected considering both scene structure and radiance for such regions. 

The examples in \fref{fig:compDenoising} and \fref{fig:ResultsDemosaickingDefogging2} obviously show that the proposed algorithm suppresses noise amplification effectively without degrading defogging performance. Indeed, noise effects are clearly visible in distant scenes and denser fog condition as analyzed in \sref{sec:AnalysisNoise}. The proposed TLS approach shows the benefits in the environments with noise since it estimates missing values with the aid of neighbor information~\cite{Hirakawa06}. We also compared our results to the denoised output images by the state-of-the-art denoising algorithms: non-local mean~(NLM)~\cite{Buades05}, block-matching and 3D filtering~(BM3D)~\cite{Dabov07}, and expected patch log likelihood~(EPLL)~\cite{Zoran11}. The denoised output images were processed by the order of a typical pipeline--demosaicking, denoising, and defogging. The examples in \fref{fig:compDenoising} and \fref{fig:Denoising2} show that the proposed method solves demosaicking, denoising, and defogging problems in a step with minimal post-defogging artifacts. The regions of interest in \fref{fig:compDenoising} after applying the denoising algorithms are displayed in \fref{fig:jointROI}. These examples imply that denoising is not able to alleviate demosaicking artifacts and may produce additional uncertainty in the subsequent defogging process.

\section{Conclusion}\label{sec:conclusion}
It is necessary to improve the perceptual quality of images acquired in bad weather conditions before using them in many computer vision applications. Defogging technique is one of image enhancement methods commonly applied as a post-processing in a digital camera processing pipeline. Since the performance of defogging algorithm is strongly dependent on the artifacts generated by demosaicking and the physical nature of acquisition noise, we have investigated on how defogging algorithms interact in a digital camera processing pipeline in this paper. These analysis and observations demonstrate that   demosaicking artifacts are exacerbated by the subsequent defogging process and the artifacts are unremovable by denoising process. The proposed joint defogging and demosaicking algorithm proved that solving the defogging and demosaicking problems jointly is more effective than treating them separately and more advantageous for suppressing noise. Lastly, we have presented subjective and objective experimental results on the real-world dataset acquired in a raw format.

\newpage
\ifCLASSOPTIONcaptionsoff
  \newpage
\fi

\bibliographystyle{IEEEtran}
\bibliography{main}

\end{document}